\pdfoutput=1

\documentclass[letterpaper]{article} 
\usepackage{aaai24}  
\usepackage{times}  
\usepackage{helvet}  
\usepackage{courier}  
\usepackage[hyphens]{url}  
\usepackage{graphicx} 
\usepackage{natbib}  
\usepackage{caption} 
\usepackage{algorithm}
\usepackage{algorithmic}

\usepackage{newfloat}
\usepackage{listings}
\DeclareCaptionStyle{ruled}{labelfont=normalfont,labelsep=colon,strut=off} 
\floatstyle{ruled}
\newfloat{listing}{tb}{lst}{}
\floatname{listing}{Listing}
\title{Learning to Unlearn: Instance-wise Unlearning for Pre-trained Classifiers}
\author{
    Sungmin Cha\textsuperscript{\rm 1}\equalcontrib,
    Sungjun Cho\textsuperscript{\rm 2}\equalcontrib, 
    Dasol Hwang\textsuperscript{\rm 2}\equalcontrib,\\
    Honglak Lee\textsuperscript{\rm 2}, 
    Taesup Moon\textsuperscript{\rm 3} and  
    Moontae Lee\textsuperscript{\rm 2,4}
}
\affiliations{
    \textsuperscript{\rm 1}New York University \
    \textsuperscript{\rm 2}LG AI Research \
    \textsuperscript{\rm 3}ASRI / INMC / Seoul National University \
    \textsuperscript{\rm 4}University of Illinois Chicago\\
    sungmin.cha@nyu.edu, sungjun.cho@lgresearch.ai, dasol.hwang@lgresearch.ai,\\ honglak@lgresearch.ai, tsmoon@snu.ac.kr, moontae.lee@lgresearch.ai
}

\usepackage{amsmath}
\usepackage{amssymb}
\usepackage{booktabs}
\usepackage{bm}
\usepackage{multirow}
\usepackage{subfigure}
\usepackage{pifont}
\usepackage{xcolor}

\newcommand{\cmark}{\ding{51}}%
\newcommand{\xmark}{\ding{55}}%

\DeclareMathOperator*{\argmin}{arg\,min}

\begin{document}

\maketitle
\begin{abstract}
Since the recent advent of regulations for data protection (e.g., the General Data Protection Regulation), there has been increasing demand in deleting information learned from sensitive data in pre-trained models without retraining from scratch. The inherent vulnerability of neural networks towards adversarial attacks and unfairness also calls for a robust method to remove or correct information in an instance-wise fashion, while retaining the predictive performance across remaining data. To this end, we consider {\it instance-wise unlearning}, of which the goal is to delete information on a set of instances from a pre-trained model, by either misclassifying each instance away from its original prediction or relabeling the instance to a different label. We also propose two methods that reduce forgetting on the remaining data: 1) utilizing adversarial examples to overcome forgetting at the representation-level and 2) leveraging weight importance metrics to pinpoint network parameters guilty of propagating unwanted information. Both methods only require the pre-trained model and data instances to forget, allowing painless application to real-life settings where the entire training set is unavailable. Through extensive experimentation on various image classification benchmarks, we show that our approach effectively preserves knowledge of remaining data while unlearning given instances in both single-task and continual unlearning scenarios.
\end{abstract}

\section{Introduction}

Humans remember and forget: efficiently learning useful knowledge yet regulating privately sensitive information and protecting from malicious attacks. Recent advances in large-scale pre-training enable models to memorize massive information for intelligent operations~\cite{(gpt2)radford2019language}, but this strength comes at a cost. In computer vision, models trained on numerous image data often misclassify naturally adversarial or adversarially attacked examples with high-confidence~\cite{hendrycks2021natural}. In natural language processing, models trained on indiscriminately collected data often disclose private information such as occupations, phone numbers, and family background during text generation~\cite{privacyllm}. 

A na\"ive solution is to retrain these models from scratch after refining or reweighting their training datasets~\cite{lison2021anonymisation,zemel2013learning,lahoti2020fairness}. However, such post-hoc processing is impractical due to growing volumes of data and substantial cost of large-scale training: while exercising the Right to be Forgotten~\cite{(rightforgotten)rosen2011right, villaronga2018humans} may be straightforward to humans, it is not so straightforward in the context of machine learning. This has sparked the field of \textit{machine unlearning}, in which the main goal is to efficiently delete information while preserving information on the remaining data.

While several machine unlearning approaches have shown promising results deleting data from traditional machine learning algorithms~\cite{linear,makingaiforget,randomforests} as well as DNN-based classifiers~\cite{fastyet,zeroshot,recoverable,fewshot,eternal,efficienttwo,hessian}, existing work are built upon assumptions far too restrictive compared to real-life scenarios. First off, many approaches assume a \textit{class-wise unlearning} setup, where the task is to delete information from all data points that belong to a particular class or set of classes. However, data deletion requests are practically received on a per-instance basis, potentially resulting in a set of data points with a mixture of class labels~\cite{privacyllm, mehrabi2021survey}. Another widely used assumption is that at least a subset of the original training data is available at the time of unlearning~\cite{fastyet,eternal,efficienttwo,hessian}. Unfortunately, loading the original dataset may not be a practical option due to data expiration policies or lack of storage for large data. Lastly, many approaches consider the main objective as \textit{undoing} the previous effect of the deleting data during training. However, previous work have shown that fulfilling this objective can still lead to information leakage~\cite{graves2021amnesiac}.

The aforementioned limitations are also reflected in a recently initialized unlearning challenge~\cite{neuripschallenge}, centered around a scenario where an age classification model trained on facial images is required to \textit{forget} specific images in order to uphold privacy rights. This marks a departure from previous approaches that primarily adopted a class-wise unlearning setup towards \textit{instance-wise unlearning}, while mirroring constraints present in real-world situations such as lack of access to the original training set.

In response, we propose a framework for instance-wise unlearning that deletes information given access only to the pre-trained model and the data points requested for deletion. Instead of undoing the previous influence of deleting data, we pursue a stronger goal where all data points requested for deletion are misclassified, preventing collection of information via interpolation of nearby data points. Achieving this stronger goal for unlearning while maintaining the efficacy of the pre-trained model on the remaining data is a challenging task. Inspired by works in continual learning and adversarial attack literature~\cite{ebrahimi2020adversarial, (MAS)aljundi2018memory}, we thus propose two regularization methods that minimize the loss in predictive performance on the remaining data.

Specifically, we 1) generate adversarial examples by attacking each deleting data point with the pre-trained model and retrain on these examples to prevent representation-level forgetting and 2) use weight importance measures from unlearning instances to focus gradient updates more towards parameters responsible for the originally correct classification of such instances.
Extensive experiments on CIFAR-10, CIFAR-100~\cite{cifar} and ImageNet-1K~\cite{imagenet} datasets show that our proposed method effectively preserves overall predictive performance, while completely forgetting images requested for deletion.
Our qualitative analyses also reveal interesting insights, including lack of any discernible pattern in misclassification that may be exploited by adversaries, preservation of the previously learned decision boundary, and forgetting of high-level features within deleted images. 
In summary, our \textbf{main contributions} are as follows:
\begin{itemize}
    \item We propose \textit{instance-wise unlearning} through intended misclassification, which can effectively unlearn data with only the pre-trained model and data to forget.
    \item We present two model-agnostic regularization methods that reduce forgetting on the remaining data while misclassifying data requested for deletion.
    \item Empirical evaluations on well-known image classification benchmarks show that our proposed method significantly boosts predictive performance after unlearning.
\end{itemize}

\section{Related Work}

\noindent\textbf{Machine unlearning.}\ \
The main goal of machine unlearning~\cite{cao2015} is to adjust a pre-trained model to forget information learned from a specified subset of data. The first set of studies have proposed approaches to delete the influence of unwanted data points from shallow machine learning models such as linear/logistic regression, k-means clustering, and random forests, while retaining the predictive performance on remaining data ~\cite{linear,makingaiforget,randomforests}.

Recent work have studied machine unlearning for deep neural networks in various setups, which we summarize in Table~\ref{tab:comparison}. These methods can be categorized into two approaches: \textit{class-wise} and \textit{instance-wise} unlearning. Class-wise unlearning forgets all data points belonging to a certain class (\textit{e.g.,} all images of dogs in CIFAR-10) while retaining performance on the remaining classes~\cite{fastyet,zeroshot,recoverable,fewshot, graves2021amnesiac}. In contrast, instance-wise unlearning deletes information from individual data points, possibly with mixed classes~\cite{eternal,efficienttwo,hessian}.

\begin{table}[!t] 
\centering
\resizebox{\linewidth}{!}{
\footnotesize 
\begin{tabular}{l|llccc}
    \toprule
    Methods & Unit & Goal & $\mathcal{D}_r$ & $\mathcal{D}_f$ & Updates \\
    \midrule
    \midrule
    \citet{fastyet} &  class & undo & \cmark	& \xmark & \xmark \\  
    \citet{graves2021amnesiac} & class & undo &	\xmark & \xmark& \cmark \\  
    \citet{zeroshot} & class & undo & \xmark	& \xmark& \xmark \\  
    \citet{recoverable}& class & undo & \xmark& \cmark & \xmark \\  
    \citet{fewshot} & class & undo &	\xmark & \cmark& \xmark \\  
    \midrule
    \citet{eternal} & instance & undo & \cmark & \cmark& \xmark \\  
    \citet{efficienttwo} & instance & undo & \cmark & \cmark & \xmark \\  
    \citet{hessian} & instance & undo &\cmark & \cmark& \xmark  \\ 
    \midrule
    \textbf{Ours} & \textbf{instance} & \textbf{misclassify} & \xmark & \cmark & \xmark  \\
    \bottomrule
\end{tabular}
}
\caption{Comparison between existing unlearning methods.}\label{tab:comparison}
\end{table}

Previously, all methods have set the unlearning objective of guiding the pre-trained model towards the model retrained on the dataset after removing unwanted data instances (\textit{i.e.} to \textit{undo} their effect during training). Unfortunately, previous work have shown that this goal fails to provide reliable security against data leakage~\cite{graves2021amnesiac}, due to high interpolation capability of deep neural networks. Therefore, we instead set our goal for unlearning as tuning the pre-trained model towards intentionally \textit{misclassify} the data points requested for deletion. Furthermore, existing methods assume different access levels to the unlearning data $\mathcal{D}_f$ and the remaining data $\mathcal{D}_r$. For our proposed methods, we require access to no more than the unlearning data $\mathcal{D}_f$, allowing seamless application for practitioners under real-world scenarios.

\noindent\textbf{Adversarial examples.}\ \
Since the vulnerability of neural networks has been revealed~\cite{szegedy2013intriguing}, various methods have been proposed to generate adversarial examples that can deceive neural networks~\cite{(FGSM)goodfellow2014explaining, (BIM)kurakin2016adversarial, (PGD)madry2017towards, (C&W)carlini2017towards}.
Interestingly, \citet{ilyas2019adversarial} have experimentally demonstrated that adversarial examples contain useful features about the attack target label for the model. From this observation, we leverage adversarial examples as a means for regularizing the model to preserve its previously learned decision boundary in the feature space.

\begin{figure*}[!t]
    \centering
    \includegraphics[width=0.8\textwidth]{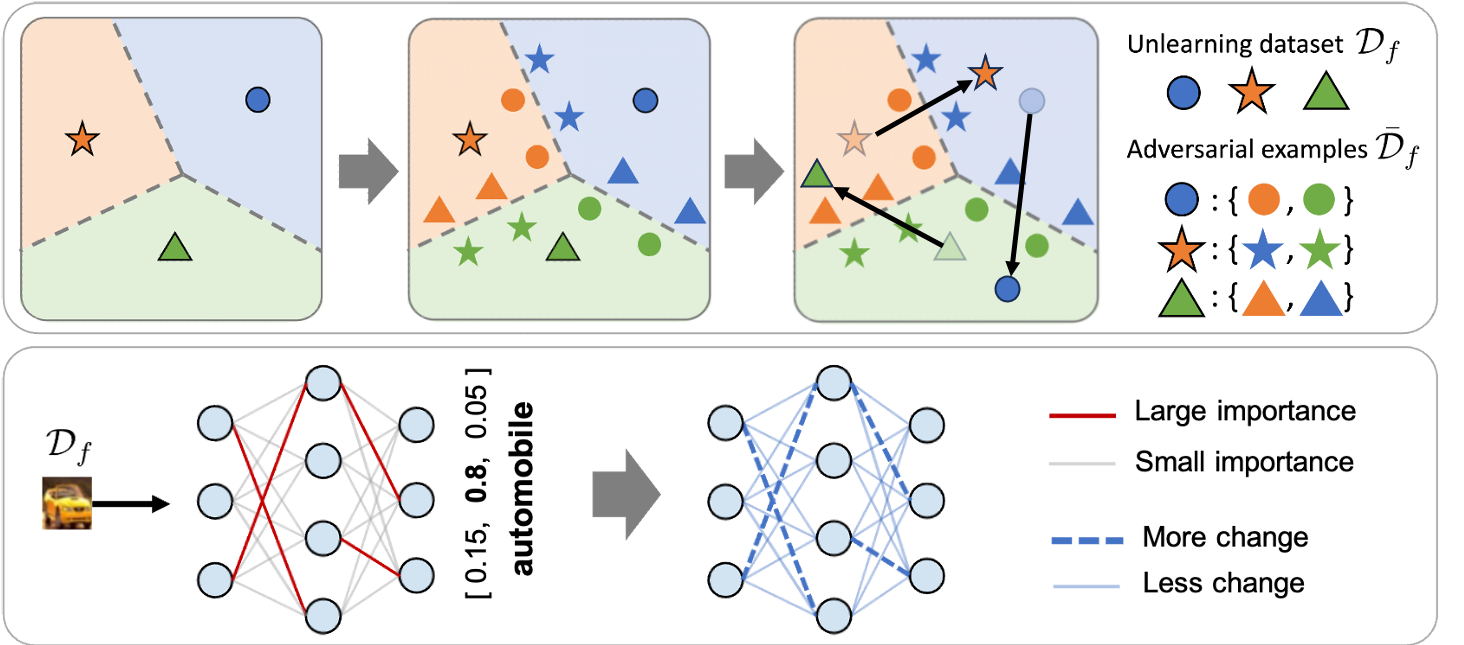}
    \caption{Illustrations of our approaches that reduce forgetting on the remaining data. (Top) Augmenting adversarial examples from unlearning data provides support for preserving the overall decision boundary. (Bottom) Weight importance measures allow us to pinpoint weights we should change to induce misclassification while maintaining other weights to mitigate forgetting.}
    \label{fig:method}
\end{figure*}

\noindent\textbf{Weight importance.}\ \
Weight importance measures how important each parameter is when the model predicts an output for a given input data, and has been used for various purposes such as model pruning~\cite{(pruning1)molchanov2019importance,(pruning2)liu2017learning,(pruning3)wen2016learning,(pruning4)alvarez2016learning,(pruning5)li2016pruning} and regularization-based continual learning~\cite{(ewc)kirkpatrick2017overcoming,(MAS)aljundi2018memory,(Rwalk)chaudhry2018riemannian,(AGS-CL)jung2020adaptive, (selfless)aljundi2018selfless, (cpr)chacpr}.
In the case of regularization-based continual learning, the weight-level importance is utilized as the strength of the L2 regularization between a current model's weight and the model's weight trained up to the previous task for overcoming catastrophic forgetting of previous tasks~\cite{(ewc)kirkpatrick2017overcoming,(MAS)aljundi2018memory}.

\section{Method}
\subsection{Preliminaries and notations}

\noindent\textbf{Dataset and pre-trained model.}\ \
Let $\mathcal{D}_{train}$ be the entire training dataset used to pre-train a classification model $g_{\bm \theta}:\mathcal{X}\rightarrow \mathcal{Y}$. We denote $\mathcal{D}_f \subset \mathcal{D}_{train}$ as the unlearning dataset that we want to intentionally forget from the pre-trained model and $\mathcal{D}_r$ as the remaining dataset on which we wish to maintain predictive accuracy ($\mathcal{D}_r := \mathcal{D}_{train} \setminus \mathcal{D}_f$). We denote a pair of an input image $\bm x \in \mathcal{X}$ and its ground-truth label $\bm y \in \mathcal{Y}$ from $\mathcal{D}_{train}$ as $(\bm x, \bm y) \sim \mathcal{D}_{train}$, similarly $(\bm x_f, \bm y_f) \sim \mathcal{D}_{f}$ and $(\bm x_r, \bm y_r) \sim \mathcal{D}_{r}$. Also, $\mathcal{D}_{test}$ denotes the test dataset used for evaluation. Note that our approaches assumes access to only the pre-trained model $g_{\theta}$ and the unlearning dataset $\mathcal{D}_f$ during unlearning.

\noindent\textbf{Adversarial examples.}\ \
The goal of an adversarial attack on an input $(\bm x, \bm y)$ is to generate an adversarial example $\bm {x'}$ that is similar to $\bm x$, but leads to misclassification ($g_{\bm \theta}(\bm {x'})\neq \bm y$) when fed to the pre-trained model $g_{\bm {\theta}}$. In the case of \textit{targeted} adversarial attack, it makes the model predict a specific class different from the true class ($g_{\bm \theta}(\bm {x'}) = \bar{\bm y}$). The typical optimization form of generating adversarial examples in targeted attack is denoted as
\begin{equation}\label{eq:adversarial_attack}
    \bm x' = \argmin_{\bm z:\|\bm z - \bm x\|_{p} \le \epsilon} \mathcal{L}_{\textrm{CE}}(g_{\bm {\theta}}(\bm z), \bar{\bm y} ; \bm {\theta})
\end{equation}
where $\mathcal{L}_{\textrm{CE}}$ stands for the cross-entropy loss. The $\|\bm z- \bm x\|_{p} \leq \epsilon$ condition requires that the $L_p$-norm is less than a perturbation budget $\epsilon$. The optimization above is intractable in general, and there exist approximations that can generate adversarial examples without directly solving it~\cite{(FGSM)goodfellow2014explaining,(BIM)kurakin2016adversarial,(C&W)carlini2017towards,(PGD)madry2017towards}. For our experiments, we make use of $L_2$-PGD targeted attacks~\cite{(PGD)madry2017towards}.

\noindent\textbf{Measuring weight importance with MAS.}\  \
To measure weight importance $\Omega$, we consider MAS~\cite{(MAS)aljundi2018memory}, an algorithm that estimates weight importance by finding parameters that bring a significant change in the output when perturbed slightly. It estimates the weight importance via a sum of gradients on the L2-norm of the outputs:
\begin{equation}\label{eq:mas}
\Omega_{i} = \frac{1}{N} \sum_{n=1}^N {\bigg |{\frac{\partial \| g_{\bm {\theta}}(\bm x^{(n)}; \bm \theta)\|^{2}_{2}}{\partial \theta_{i}}}\bigg |}
\end{equation}
where $i$ denotes the index of model parameter weights and $\bm x^{(n)}$ denotes the $n$-th input image from a total of $N$ numbers of images. Note that each $\Omega_{i}$ can be interpreted as a measure of importance of $\theta_{i}$ in producing the output of $N$ images.

\subsection{Our proposed framework}\label{sec:method}
\noindent\textbf{Instance-wise unlearning.} \ \
Let $\hat{g}_{\bm {\theta}}$ denote the model after unlearning.
We consider two types of misclassification for instance-wise unlearning:
(\textbf{\lowercase\expandafter{\romannumeral1}}) \textit{misclassifying} all data points in $\mathcal{D}_f$, (\textit{i.e.,} $\hat{g}_{\bm {\theta}}(\bm {x}_f) \neq \bm {y}_f$).
(\textbf{\lowercase\expandafter{\romannumeral2}}) \textit{relabeling (or correcting)} the predictions of $\mathcal{D}_f$ (\textit{i.e.,} $\hat{g}_{\bm {\theta}}(\bm {x}_f) = \bm {y}_{f}^{*}$) where $\bm {y}_{f}^{*} \neq \bm {y}_{f}$ is chosen individually for each input $\bm {x}_f$. Let $\mathcal{L}_\textrm{UL}$ denote the unlearning loss function used to train a classification model.
The two goals above can be realized by minimizing the following loss functions:
\begin{equation}\label{eq:ul_mis}
  \mathcal{L}_{\textrm{UL}}^{\textrm{MS}}(\mathcal{D}_f; \bm \theta) = -\mathcal{L}_{\textrm{CE}}(g_{\bm {\theta}}(\bm {x}_f), \bm {y}_f; \bm \theta)
\end{equation}
\begin{equation}\label{eq:ul_correction}
 \mathcal{L}_{\textrm{UL}}^{\textrm{Cor}}(\mathcal{D}_f; \bm \theta) = \mathcal{L}_{\textrm{CE}}(g_{\bm {\theta}}(\bm {x}_f), \bm {y}_{f}^{*}; \bm \theta)
\end{equation}

When unlearning solely based on the two loss functions above, the model is likely to suffer from significant \textit{forgetting} on $\mathcal{D}_{r}$. Thus, a crucial objective shared across both unlearning goals is to prevent forgetting of previously acquired knowledge and classify data in $\mathcal{D}_r$ as accurately as possible.

When both $\mathcal{D}_f$ and $\mathcal{D}_r$ are available, we can easily obtain an \textit{oracle} model that satisfies the objective by re-training the model with the following loss function:  $\mathcal{L}_{\textrm{oracle}}(\mathcal{D}_f,\mathcal{D}_r; \bm \theta) = \mathcal{L}_\textrm{UL}(\mathcal{D}_f; \bm \theta) +\mathcal{L}_{\textrm{CE}}(\mathcal{D}_r; \bm \theta)$. However in real-settings, access to $\mathcal{D}_r$ may not be an option due to high cost in data storage. To tackle this limitation, we define a regularization-based unlearning that achieves the goal above without use of $\mathcal{D}_r$:
\begin{equation}\label{eq:oracle}
\mathcal{L}_{\textrm{RegUL}}(\mathcal{D}_f; \bm \theta) = \mathcal{L}_\textrm{UL}(\mathcal{D}_f; \bm \theta) +\mathcal{R}(\mathcal{D}_f, {g}_{\bm {\theta}})
\end{equation}
Here, $\mathcal{R}(\cdot)$ is the regularization term used to overcome forgetting of knowledge on the remaining data $\mathcal{D}_r$. In the following subsections, we introduce two novel regularization methods designed to overcome representation- and weight-level forgetting during the unlearning process.

\noindent\textbf{Regularization using adversarial examples.}\  \
The motivation of using adversarial examples stems from the work of~\citet{ilyas2019adversarial}, which showed that perturbations added to $\bm x$ to generate an adversarial example $\bm x'$ contain class-specific features of the attack target label $\bar{\bm y} \neq \bm y$. Based on this finding, we utilize generated adversarial examples as part of regularization $\mathcal{R}(\cdot)$ to preserve previously learned class-specific knowledge, overcoming forgetting at the representation-level (Algorithm 1 of the supplementary materials). Let $\mathcal{D}_f$ be a set of $N_f$ images: $\{(\bm x^{(i)}_f, \bm y^{(i)}_f)\}_{i=1}^{N_f}$. Prior to unlearning, we generate adversarial examples $\bm x'_{f}$ using the targeted PGD attack with a randomly selected attack target label $\bar{\bm y} \neq \bm y_f$. We generate $N_{\textrm{adv}}$ adversarial examples per input $\bm x_{f}$, resulting in a set of $\bar{N}_f = N_f \times N_{\textrm{adv}}$ examples, denoted as $\bar{\mathcal{D}}_f = \{(\bm {x}'^{(i)}_f, \bm{ \bar{y}}^{(i)}_f)\}_{i=1}^{\bar{N}_f}$. During unlearning, we add $\mathcal{L}_{\textrm{CE}}(\bar {\mathcal{D}}_f; \bm \theta)$ with adversarial examples as regularization:
\begin{equation}\label{eq:ul_adv}
\begin{aligned}
\mathcal{L}_{\textrm{UL}}^{\textrm{Adv}}(\mathcal{D}_f; \bm \theta) &= \mathcal{L}_\textrm{UL}(\mathcal{D}_f; \bm \theta) +\mathcal{R_{\textrm{Adv}}}(\mathcal{D}_f, {g}_{\bm {\theta}}) \\
&= \mathcal{L}_\textrm{UL}(\mathcal{D}_f; \bm \theta) + \mathcal{L}_{\textrm{CE}}(\bar {\mathcal{D}}_f; \bm \theta)
\end{aligned}
\end{equation}

An intuitive illustration of this approach in the representation-level is shown in Figure~\ref{fig:method}. In essence, the generated adversarial examples in $\bar {\mathcal{D}}_f$ mimic the remaining dataset $\mathcal{D}_r$, providing information of the pre-trained decision boundary within the representation space. As a result, by adding $\mathcal{L}_{\textrm{CE}}(\bar {\mathcal{D}}_f; \bm \theta)$ as a regularizer to the unlearning process, the model learns a new decision boundary that minimizes $\mathcal{L}_\textrm{UL}$ (in Eq.~\ref{eq:ul_mis} and \ref{eq:ul_correction}) while simultaneously attempting to keep the decision boundary of the original model.

\noindent\textbf{Regularization with weight importance.}   \ \
We also propose a regularization using weight importance to overcome forgetting at the weight-level. As depicted in Figure~\ref{fig:method}, our approach is to maintain the weights that are less important for predictions on $\mathcal{D}_f$ as much as possible, while allowing changes in weights that are considered important for correctly predicting $\mathcal{D}_f$. In other words, we penalize changes of weights that were less important when classifying $\mathcal{D}_f$, thereby preventing weight-level forgetting.

\begin{table*}[!tp]
\footnotesize
\centering
\smallskip\noindent
\resizebox{.9\linewidth}{!}{
\begin{tabular}{ll|cccc|cccc|cccc}
\toprule
&  & \multicolumn{4}{c|}{CIFAR-10} & \multicolumn{4}{c|}{CIFAR-100} & \multicolumn{4}{c}{ImageNet-1K}\\ 
&  & {$k=16$} & {$k=64$} & $k=128$ & $k=256$  & {$k=16$} & {$k=64$} & $k=128$ & $k=256$    & {$k=16$} & {$k=64$} & $k=128$ & $k=256$   \\ \midrule
\multirow{6}{*}{$\mathcal{D}_f$ ($\downarrow$)} 
& \textsc{Before} & {100.0} & {99.38} & 99.53 & 99.38 & {100.0} & {100.0} & 100.0 & 100.0 & {87.50} & {84.69} & 86.09 & 86.02 \\
& \textsc{Oracle}  &  0.0 &  0.0 &  0.0 & 0.63 &  0.0 & 0.0 & 0.0 & 0.0 &  0.0 & 0.0  & 0.26  & 5.73   \\ 
& \textsc{NegGrad}  & 0.0   & 0.0   & 0.0   & 3.83   & 0.0   & 0.0   & 0.0   & 0.0   & 0.0   & 0.0   & 0.0   & 0.0   \\ 
& \textsc{RAWP} & 0.0   & 0.0   & 4.06   & 6.17   & 0.0   & 0.0   & 0.0   & 0.0   & 0.0   & 0.0   & 0.0   & 0.0    \\ 
& \textsc{Adv}  & 0.0   & 0.0   & 0.0   & 0.0   & 0.0   & 0.0   & 0.0   & 0.0   & 0.0   & 0.0   & 0.0   & 0.0   \\ 
& \textsc{Adv+Imp}  & 0.0   & 0.0   & 0.0   & 0.0   & 0.0   & 0.0   & 0.0   & 0.0   & 0.0   & 0.0   & 0.0   & 2.73\\ 
\midrule
\multirow{6}{*}{$\mathcal{D}_r$ ($\uparrow$)}    
& \textsc{Before} & {99.60} & {99.60} & 99.60 & 99.60 & {99.98} & {99.98} & 99.98 & 99.98 & {87.42} & {87.42} & 87.42 & 87.42\\
& \textsc{Oracle}  & 98.74  & 99.72  & 98.97 & 39.90 & 99.96 & 96.17  & 96.74 & 96.43 &  73.36 & 77.81 & 67.7 &  61.61 \\
& \textsc{NegGrad}  & {15.79} & {9.22}  & 7.11  & 6.34 & {66.97} & {26.20} & 11.64 & 1.70 & {65.02} & {35.66} & 19.27 & 0.48\\
& \textsc{RAWP}  & 67.15  & 37.32 & 13.28 & 9.12 & 72.44  & 17.39 & 2.82 & 1.12   & 20.33 & 2.72 & 0.29 & 0.10  \\ 
& \textsc{Adv}  & {69.70} & {66.97} & 53.49 & 39.33 & {89.18} & {81.07} & 76.28 & 65.67 & {80.92} & {73.60} & 66.31 & 43.70\\
& \textsc{Adv+Imp}  & {\textbf{85.75}} & {\textbf{72.77}} & \textbf{54.51} & \textbf{48.95} & {\textbf{89.91}} & {\textbf{89.48}} & \textbf{82.86} & \textbf{67.60} & {\textbf{81.43}} & {\textbf{79.03}} & \textbf{74.14} & \textbf{65.32}\\ 
\midrule
\multirow{6}{*}{$\mathcal{D}_{test}$ ($\uparrow$)} 
& \textsc{Before} & {92.59} & {92.59} & 92.59 & 92.59 & {77.10} & {77.10} & 77.10 & 77.10  & {76.01} & {76.01} & 76.01 & 76.01\\
& \textsc{Oracle}  & 90.21 &  91.01 &  89.44 & 38.59 & 64.41 & 67.06 & 66.88 & 65.31 & 63.91 & 68.08 & 59.09 &  54.03 \\ 
& \textsc{NegGrad} & {15.87} & {9.28}  & 7.11  & 6.47 & {48.07} & {21.11} & 10.19 &  1.71 & {56.70} & {31.10} & 17.08 & 0.46\\
& \textsc{RAWP}  & 63.10 & 35.47 & 12.91 & 9.12 & 50.99 & 14.41  & 2.79 & 1.12 & 17.53 & 2.41 & 0.28 & 0.10  \\ 
& \textsc{Adv}  & {65.14} & {62.23} & 49.47 & 36.69 & {63.17} & {57.43} & 53.89 & 46.45 & {70.36} & {65.51} & 57.34 & 38.25\\
& \textsc{Adv+Imp}  & {\textbf{79.65}} & {\textbf{67.08}} & \textbf{50.82} & \textbf{45.44} & {\textbf{63.69}} & {\textbf{62.83}} & \textbf{58.44} & \textbf{48.71} & {\textbf{70.77}} & {\textbf{68.43}} & \textbf{64.05} & \textbf{56.52}\\ 
\bottomrule
\end{tabular}}
\caption{Evaluation results before and after unlearning $k$ instances from ResNet-18 (for CIFAR-10) and ResNet-50 (for CIFAR-100 and ImageNet-1K) pretrained on respective image classification datasets. While using negative gradients only loses significant information on $\mathcal{D}_r$, our proposed methods \textsc{Adv} and \textsc{Adv+Imp} retain predictive performance on $\mathcal{D}_r$ as well as $\mathcal{D}_{test}$, while completely forgetting instances in $\mathcal{D}_f$.}\label{table:main_result_dataset}
\end{table*}

Specifically, we calculate the weight importance with MAS~\cite{(MAS)aljundi2018memory} given $g_{\bm \theta}$ and $\mathcal{D}_f$ before unlearning, and normalize the importance measurements $\Omega^l$ within each $l$-th layer to lie within $[0,1]$. Note that this normalized importance $\Omega^l$ assigns large values to weights important for classifying $\mathcal{D}_f$. Therefore, we add a regularization that sums all parameter change in the $l$-th layer weighted by $\bar{\Omega}^l = 1 - \Omega^l$, such that more important weights are updated more. The objective including weight importance with regularization via adversarial examples can be written as:
\begin{equation}\label{eq:ul_adv_imp}
\begin{aligned}
\mathcal{L}_{\textrm{UL}}^{\textrm{Adv+Imp}}(\mathcal{D}_f; \bm \theta)& = \mathcal{L}_{\textrm{UL}}^{\textrm{Adv}}(\mathcal{D}_f; \bm \theta) +\mathcal{R_{\textrm{Imp}}}(\mathcal{D}_f, {g}_{\bm {\theta}}) \\
& = \mathcal{L}_{\textrm{UL}}^{\textrm{Adv}}(\mathcal{D}_f; \bm \theta) + \sum_{i}^{} \bar{\Omega}_i(\theta_{i} - \Tilde{\theta}_{i})^2
\end{aligned}
\end{equation}
where $i$ is the index of each weight and $\Tilde{\theta}$ is the initial weight of the pre-trained classifier before unlearning. The pseudocode of measuring weight importance is shown in Algorithm 2 of the supplementary material. Through various experiments, we observe that applying the regularization using adversarial examples is already effective to overcome forgetting of knowledge in $\mathcal{D}_r$, and the additional regularization with weight importance further enhances performance even further, especially in more challenging scenarios such as continual unlearning. The pseudocode of our overall unlearning pipeline is presented in the supplementary material.

\section{Experiments}

\subsection{Experimental Setup}

\noindent\textbf{Datasets and baselines.} \ \
We evaluate our unlearning methods on three different image classification datasets: CIFAR-10, CIFAR-100~\cite{cifar}, and ImageNet-1K~\cite{imagenet}. 
We use ResNet-18 as the base model for CIFAR-10, and ResNet-50 for CIFAR-100 and ImageNet-1K~\cite{(resnet)he2016deep}.
Experimental results from other base models such as MobileNetv2~\cite{(mobilenetv2)sandler2018mobilenetv2}, SqueezeNet~\cite{(squeezenet)iandola2016squeezenet}, DenseNet~\cite{(densenet)huang2017densely}, and ViT~\cite{(vit)dosovitskiy2020image} are also available in the supplementary material.

The compared baselines are as follows: \textsc{Before} corresponds to the pre-trained model before unlearning; \textsc{Oracle} denotes the ideal model that is retrained with positive gradients from $\mathcal{D}_r$ and negative gradients from $\mathcal{D}_f$ until maximum performance is achieved on $\mathcal{D}_r$ (\textit{i.e.}, $\mathcal{L}_{\textrm{oracle}}$). In case the accuracy of $\mathcal{D}_f$ does not reach 0 after a certain number of epochs, we report results based on the checkpoint that attained the lowest performance on $\mathcal{D}_f$; \textsc{NegGrad}~\cite{eternal} fine-tunes the model on $\mathcal{D}_f$ using negative gradients (\textit{i.e.}, $\mathcal{L}_{\textrm{UL}}^{\textrm{MS}}$) without any regularization; Repeated adversarial weight perturbation (\textsc{RAWP}) is another baseline derived from AWP~\cite{(awp)wu2020adversarial} that induces misclassification by adding weight perturbations generated from $\mathcal{D}_f$ to the model parameters. \textsc{Adv} denotes our proposed method using adversarial examples (\textit{i.e.}, $\mathcal{L}_{\textrm{UL}}^{\textrm{Adv}}$); Lastly, \textsc{Adv+Imp} uses adversarial examples as well as weight importance for regularization (\textit{i.e.}, $\mathcal{L}_{\textrm{UL}}^{\textrm{Adv+Imp}}$).

\noindent\textbf{Experimental details.} \ \
For each dataset, we randomly pick $k \in \{16, 64, 128, 256\}$ images from the entire training dataset as unlearning data $\mathcal{D}_f$ and consider the remaining data as $\mathcal{D}_r$. For unlearning, we use a SGD optimizer with a learning rate of 1e-3, weight decay of 1e-5, and momentum of 0.9 across all experiments. We early-stop training when the model attains 0\% or 100\% accuracy on the unlearning data $\mathcal{D}_f$, in case of \textit{misclassifying} and \textit{relabeling}, respectively. For generating adversarial examples from $\mathcal{D}_f$, we use $L_2$-PGD targeted attack~\cite{(PGD)madry2017towards} with a learning rate of 1e-1, attack iterations of 100 and $\epsilon=0.4$. It generates 20 adversarial examples per image for CIFAR-10 and 200 examples per image for CIFAR-100 and ImageNet-1K. For the weight importance regularization,  we set regularization strength $\lambda = 1$ in Eq.~\ref{eq:oracle}.

\subsection{Main Results}

\noindent\textbf{Results on various datasets.}  \ \
Table~\ref{table:main_result_dataset} shows results before and after unlearning (misclassifying) $k$ instances from ResNet-18 and ResNet-50 models pre-trained on CIFAR-10 and CIFAR-100/ImageNet-1K, respectively. In terms of accuracies on $\mathcal{D}_f$, we find that all methods can almost forget up to $k=256$ instances with consistently zero post-unlearning accuracy, except for some cases (\textit{e.g.}, \textsc{Oracle}, \textsc{NegGrad} and \textsc{RAWP} in CIFAR-10 when $k=256$). For most cases, \textsc{Oracle} shows not only superior accuracy in the remaining data (i.e. $\mathcal{D}_r$ and $\mathcal{D}_{test}$) but also almost zero accuracy in $\mathcal{D}_f$. Nevertheless, we find that as $k$ increases, its performance decreases, demonstrating the difficulty of achieving the proposed goal of unlearning. \textsc{NegGrad} results in a significant loss of accuracy on the remaining data for all datasets. \textsc{RAWP} achieves competitive accuracy in the remaining data than \textsc{NegGrad} in CIFAR-10. However, it performs significantly worse when the dataset becomes complex (\textit{e.g.,} CIFAR-100 and ImageNet) or when $k$ becomes larger than $16$, showing that simply perturbing the parameters cannot be a direct solution.

\begin{table}[!t]
    \centering
    \noindent
    \resizebox{\linewidth}{!}{
    \begin{tabular}{ll|ccccc}
        \toprule
        &  & \multicolumn{5}{c}{UTKFace}\\ 
        &  & $k=1$ & $k=2$  & $k=4$ & $k=8$ & $k=16$ \\ 
        \midrule
        \multirow{4}{*}{$\mathcal{D}_f$ ($\downarrow$)}
        & \textsc{Before} & 100.00 & 100.00 & 100.00 & 100.00 & 100.00\\
        & \textsc{NegGrad} & 0.00 & 0.00 & 0.00 & 0.00 & 0.00\\
        & \textsc{Adv} & 0.00 & 0.00 & 0.00 & 0.00 & 0.00\\
        & \textsc{Adv+Imp} & 0.00 & 0.00 & 0.00 & 0.00 & 0.00\\
        \midrule
        \multirow{4}{*}{$\mathcal{D}_r$ ($\uparrow$)}  
        & \textsc{Before} & 99.85 & 99.85 & 99.84 & 99.84 & 99.83\\
        & \textsc{NegGrad} & 75.75 & 36.80 & 28.95 & 28.35 & 32.75\\
        & \textsc{Adv} & 79.20 & 62.69 & 80.63 & 72.94 & 56.17\\
        & \textsc{Adv+Imp} & \bf 87.07 & \bf 63.85 & \bf 82.30 & \bf 74.33 & \bf 57.21\\
        \midrule
        \multirow{4}{*}{$\mathcal{D}_{test}$ ($\uparrow$)} 
        & \textsc{Before} & 87.14 & 87.14 & 87.14 & 87.14 & 87.14\\
        & \textsc{NegGrad} & 72.67 & 36.57 & 29.40 & 27.16 & 32.14\\
        & \textsc{Adv} & 76.30 & 58.46 & 78.28 & 65.50 & 53.65\\
        & \textsc{Adv+Imp} & \bf 82.24 & \bf 61.79 & \bf 79.54 & \bf 66.98 & \bf 54.37\\
        \bottomrule
    \end{tabular}
    }
    \caption{Instance-wise unlearning results from a ResNet-18 age classifier trained on UTKFace.}\label{table:age_prediction}
\end{table}

Meanwhile, adding regularization with adversarial examples (\textsc{Adv}) boosts the accuracy by more than 40\% depending on the number of instances to forget. 
Incorporating weight importances from MAS (\textsc{Adv+Imp}) provides further improvement, even surpassing the performance of \textsc{Oracle} in some cases (\textit{e.g.}, $k=256$ in CIFAR-10 and ImageNet).

\begin{table*}[t!]
    \begin{minipage}{0.48\textwidth}
        \footnotesize
        \centering
        \smallskip\noindent
        \resizebox{.957\linewidth}{!}{
        \begin{tabular}{ll|cccc}
            \toprule
            &  & \multicolumn{4}{c}{CIFAR-100 ($k_{CL}=8$)}\\ 
            &  & {$k=8$} & {$k=16$}  & {$k=64$} & {$k=128$} \\ 
            \midrule
            \multirow{4}{*}{$\mathcal{D}_f$ ($\downarrow$)} & \textsc{Before}  & 100.0 & 100.0 & 100.0 & 100.0\\ 
            & \textsc{NegGrad}  & 0.0 & 0.0 & 0.0 & 1.08\\ 
            & \textsc{Adv}  & 0.0 & 0.0 & 0.0 & 0.0\\ 
            & \textsc{Adv+Imp}  & 0.0 & 0.0 & 0.0 & 0.56\\  
            \midrule
            \multirow{4}{*}{$\mathcal{D}_r$ ($\uparrow$)}  & \textsc{Before}  & 99.98 & 99.98 & 99.98 & 99.98\\ 
            & \textsc{NegGrad}  & 75.68 & 30.40 & 7.00 & 1.51\\ 
            & \textsc{Adv}  & 87.02 & 71.84 & 60.12 & 39.30\\ 
            & \textsc{Adv+Imp}  & \textbf{87.42} & \textbf{72.99} & \textbf{61.76} & \textbf{48.04}\\ 
            \midrule
            \multirow{4}{*}{$\mathcal{D}_{test}$ ($\uparrow$)} & \textsc{Before}  & 77.10 & 77.10 & 77.10 & 77.10\\ 
            & \textsc{NegGrad}  & 54.51 & 24.76 & 6.22 & 0.98\\ 
            & \textsc{Adv}  & 62.76 & 51.20 & 43.40 & 31.11\\ 
            & \textsc{Adv+Imp}  & \textbf{63.01} & \textbf{52.98} & \textbf{45.55} & \textbf{36.87}\\ 
            \bottomrule
        \end{tabular}
        }
        \caption{Unlearning instances continually by increments of $k_{CL}=8$ images per step. Our methods outperform \textsc{NegGrad} in the continual unlearning scenario as well.}\label{table:continual_unlearning_results}
    \end{minipage}
    \hfill
    \begin{minipage}{0.48\textwidth}
        \footnotesize
        \centering
        \smallskip\noindent
        \resizebox{.957\linewidth}{!}{
        \begin{tabular}{ll|cccc}
            \toprule
            &  & \multicolumn{4}{c}{ImageNet-A}\\ 
            &  & {$k=16$} & {$k=32$} & {$k=64$} & $k=128$\\ 
            \midrule
            \multirow{4}{*}{$\mathcal{D}_f$ ($\uparrow$)} 
            & \textsc{Before}  & 0.0 & 0.0 & 0.0 & 0.0 \\ 
            & \textsc{Correct}  & 100.0 & 100.0 & 100.0 & 100.0 \\ 
            & \textsc{Adv}  & 100.0 & 100.0 & 95.31 & 83.44\\ 
            & \textsc{Adv+Imp}  & 100.0 & 100.0 & 10.94 & 9.38 \\  
            \midrule
            \multirow{4}{*}{$\mathcal{D}_r$ ($\uparrow$)}  
            & \textsc{Before}  & 87.46 & 87.46 & 87.46 & 87.46\\ 
            & \textsc{Correct}  & \textbf{84.41} & 83.29 & 80.79 & 77.38\\ 
            & \textsc{Adv}  & 81.75 & \textbf{83.80} & \textbf{83.74} & \textbf{83.44}\\ 
            & \textsc{Adv+Imp}  & 81.82 & 83.73 & 83.53 & 82.86\\ 
            \midrule
            \multirow{4}{*}{$\mathcal{D}_{test}$ ($\uparrow$)} 
            & \textsc{Before}  & 76.15 & 76.15 & 76.15 & 76.15\\ 
            & \textsc{Correct}  & \textbf{73.21} & 72.04 & 69.91 & 66.73\\ 
            & \textsc{Adv}  & 70.89 & \textbf{72.58} & \textbf{72.68} & \textbf{72.36}\\ 
            & \textsc{Adv+Imp}  & 70.98 & 72.51 & 72.39& 71.68\\ 
            \bottomrule
        \end{tabular}
        }
        \caption{Correcting adversarial images in ImageNet-A. \textsc{Adv} achieves the least forgetting, while \textsc{Adv+Imp} fails to correct large number of predictions due to strong regularization.}\label{table:imagenet_a_results}
    \end{minipage}
\end{table*}

For comparison with existing unlearning methods, we also conduct experiments with \textsc{Amnesiac}~\citep{graves2021amnesiac} as a baseline, details on which can be found in our supplementary material. When the data to be forgotten is small under class-wise unlearning, \textsc{Amnesiac} is effective without significantly changing the rest of the model. However, when attempting to remove multiple data points with mixed classes, its performance degrades significantly unless retrained on $\mathcal{D}_r$. Unlike our realistic unlearning setup, \textsc{Amnesiac} requires a record of parameter updates from training on sensitive data. This is viable only when sensitive data is known prior to model pre-training, which makes the method inapplicable in our framework.

Lastly, we mirror a real-world setting and empirically test our instance-wise unlearning framework on facial image data. Specifically, we first pretrain ResNet-18 on age-group classification on the UTKFace dataset consisted of 20k facial images, each labeled with the age of the subject ranging between 0 and 116~\cite{(utkface)zhifei2017cvpr}. Following previous work, we divide the age range into three ($[0,19]$, $[20,40]$, and $[40,116]$), effectively reducing the task into 3-way classification~\cite{(mfd)jung2021fair}. 
Table~\ref{table:age_prediction} shows the results. For \textsc{NegGrad}, we find that its performance degrades quickly, with its classification performance being lower than that of a random baseline (\textit{i.e.}, 33\%) when unlearning more than 4 instances. On the other hand, both \textsc{Adv} and \textsc{Adv+Imp} consistently outperform the na\"ive baseline across all $k$, showing that our regularization techniques are also effective in the facial image domain. The gap between \textsc{Adv} and \textsc{Adv+Imp} is much larger with small $k$, which demonstrates high potential of weight importance-based regularization for unlearning under a continuous stream of instance-wise data deletion requests.

\noindent\textbf{Continual unlearning.} \  \
In real-world scenarios, it is likely that data removal requests come as a stream, rather than all at once. Ultimately, despite continual unlearning requests, we need the unlearning method that can delete the requested data while maintaining performance for the rest data. Thus, we consider the setting of deleting $k=\{8, 16, 64, 128\}$ data by repeating the procedure of continually unlearning $\mathcal{D}_f$ in small fragments of size $k_{CL}=8$. Table~\ref{table:continual_unlearning_results} shows the results of continual unlearning in the model trained with ResNet-50 on CIFAR-100. We observe that \textsc{NegGrad} suffers from large forgetting as the iteration of unlearning procedure increases. On the other hand, our proposed method shows significantly less forgetting while effectively deleting for $\mathcal{D}_f$ even after multiple iterations of unlearning.

\noindent\textbf{Correcting natural adversarial examples.} \ \
Leveraging the ImageNet-A~\cite{hendrycks2021natural} dataset consisting of natural images that are misclassified with high-confidence by strong classifiers, we test whether our method can make corrections on these adversarial examples, while preserving knowledge from the original training data. For this experiment, we consider $\mathcal{D}_f$ to consist $k$ adversarial images from ImageNet-A, and adjust a ResNet-50 model pre-trained on ImageNet-1K to correctly classify $\mathcal{D}_f$ via our unlearning framework. For this setup, we employ a baseline $\textsc{Correct}$ that simply finetunes the model on $\mathcal{D}_f$ via cross-entropy loss (\textit{i.e.}, $\mathcal{L}_{\textrm{UL}}^{\textrm{Cor}}$). In Table~\ref{table:imagenet_a_results}, we find that na\"ive finetuning attains the best accuracy in both $\mathcal{D}_r$ and $\mathcal{D}_{test}$ when correcting predictions of a small number of images (e.g. $k=16$). When correcting larger number of images, however, the absence of regularization terms results in larger forgetting in $\mathcal{D}_r$ compared to \textsc{Adv} and \textsc{Adv+Imp}, with a performance gap that consistently increases with the number of adversarial images. Another takeaway is that regularization via weight importance does not help in this scenario, even showing a significant drop in $\mathcal{D}_f$ accuracy when a large number of adversarial images are introduced. This implies that using weight importances imposes too strong a regularization that correcting predictions for $\mathcal{D}_f$ itself becomes non-trivial. We conjecture that the aggregation of important parameters for predictions in $\mathcal{D}_f$ cover a large proportion of the network with large $k$, and that careful search for the Pareto optimal between accuracies on $\mathcal{D}_f$ and on $\mathcal{D}_r$ is required.

\begin{figure}[!h]
    \centering
        \subfigure[\textsc{NegGrad}]
        {\includegraphics[width=0.36\linewidth]{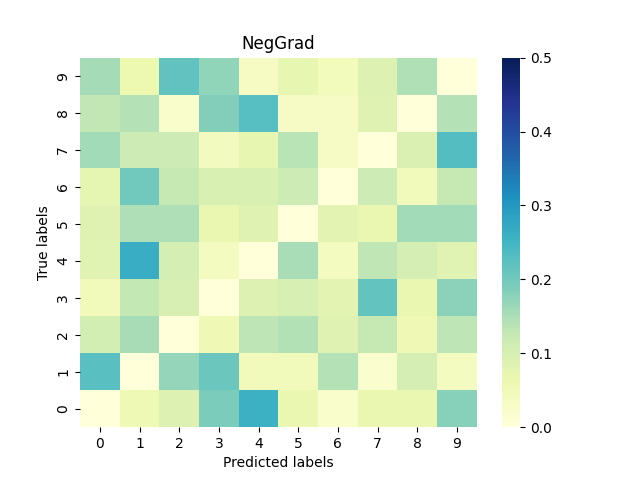}}
    \hspace{-6.0mm}
        \subfigure[\textsc{Adv}]
        {\includegraphics[width=0.36\linewidth]{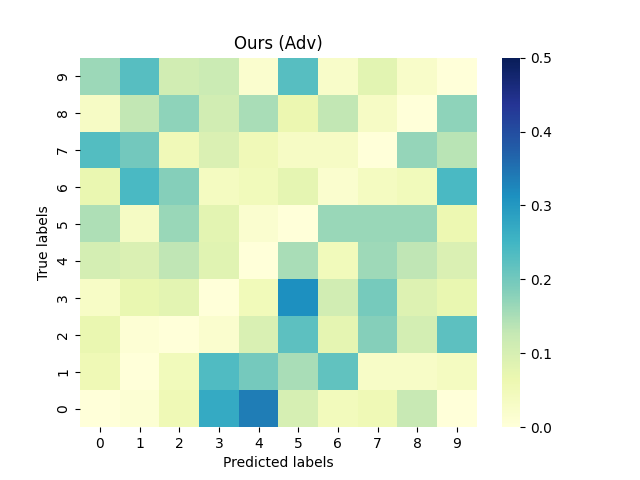}}
    \hspace{-6.0mm}
        \subfigure[\textsc{Adv+Imp}]
        {\includegraphics[width=0.36\linewidth]{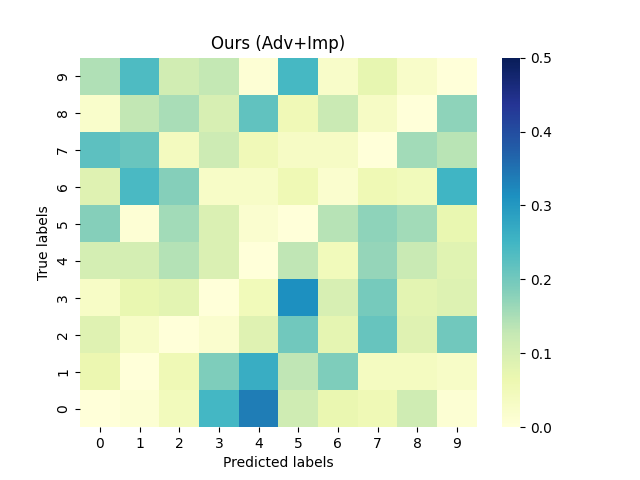}}
    \caption{Confusion matrices from CIFAR-10 showing average pairwise frequencies of pre- (Y-axis) and post-unlearning (X-axis) prediction labels from $\mathcal{D}_f$. A blue color indicates higher frequency. Our unlearning framework does not produce any discernible correlation in misclassification.}\label{fig:confusion_matrix}
\end{figure}
    \vspace{-1.0mm}

\subsection{Qualitative Analysis} 
With further analysis, we gather insight on the following questions: \textbf{Q1.} Is there any particular pattern in how the model unlearns a set of instances (\textit{i.e.,} does the model use any particular label as a retainer for deleted data)? \textbf{Q2.} How does the model isolate out instances in $\mathcal{D}_f$ from its previous decision boundary? \textbf{Q3.} How do layer-wise representations of data points in $\mathcal{D}_f$ and $\mathcal{D}_r$ change due to unlearning? For interpretable visualization, we perform the following analysis on a ResNet-18 model pre-trained on CIFAR-10.

\begin{figure}[!t]
    \centering
    \subfigure[\textsc{Before}]
    {\includegraphics[width=0.40\linewidth]{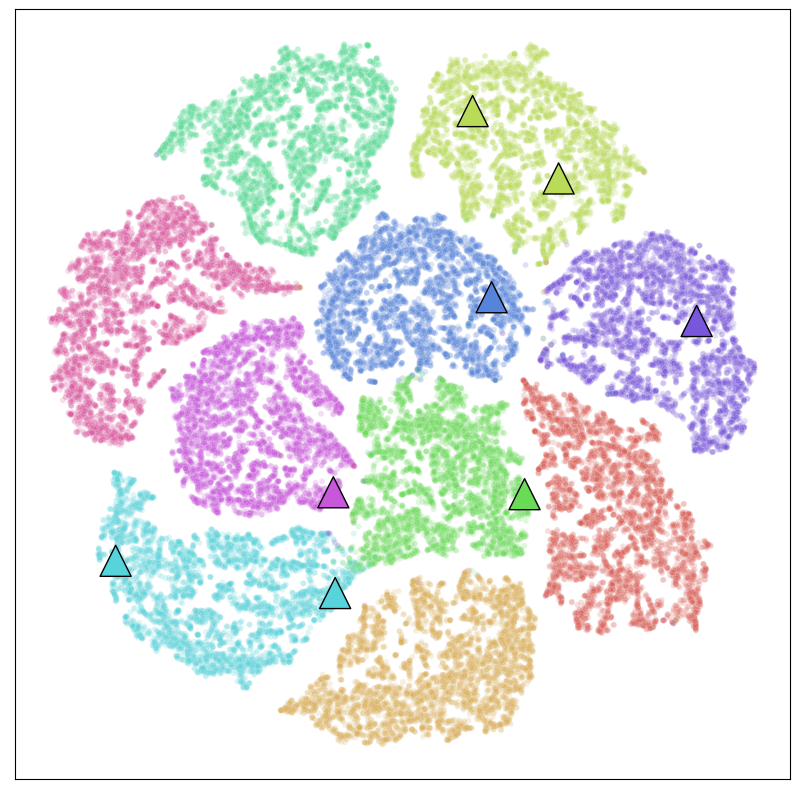}}
    \subfigure[\textsc{NegGrad}]
    {\includegraphics[width=0.40\linewidth]{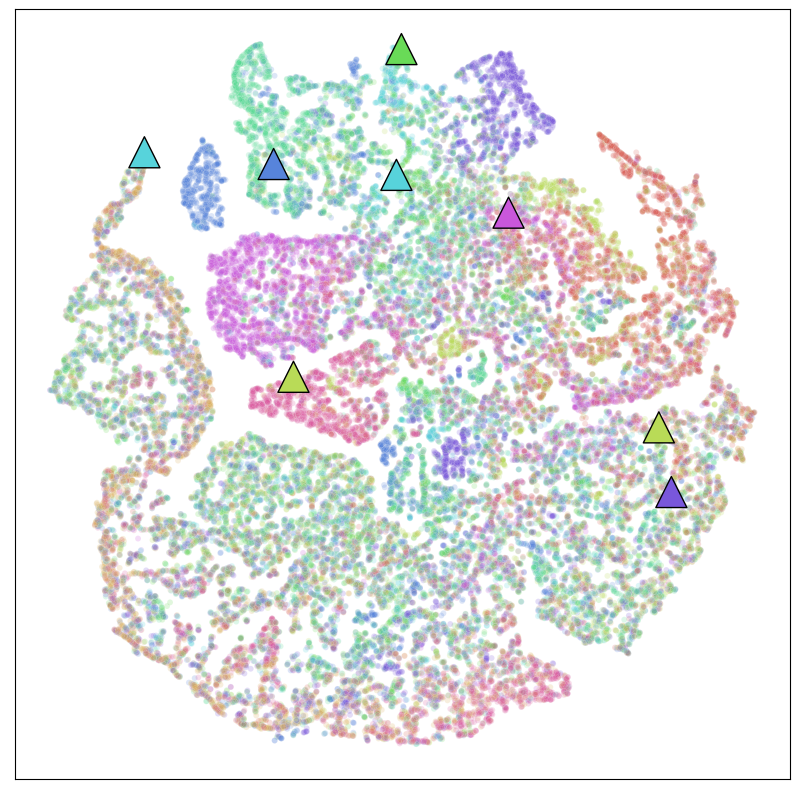}}
    \subfigure[\textsc{Adv}]
    {\includegraphics[width=0.40\linewidth]{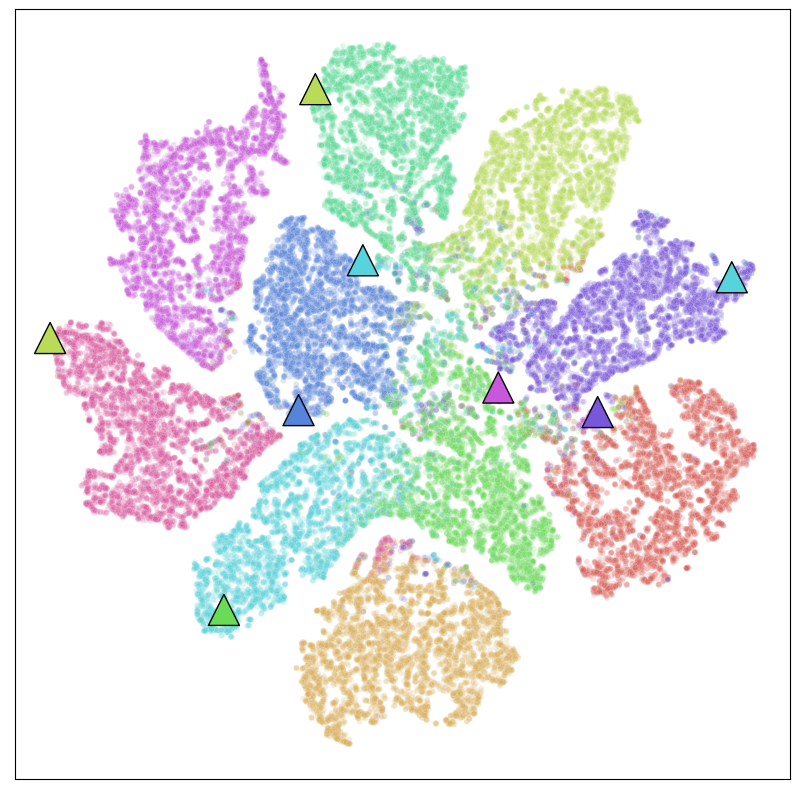}}
    \subfigure[\textsc{Adv+Imp}]
    {\includegraphics[width=0.40\linewidth]{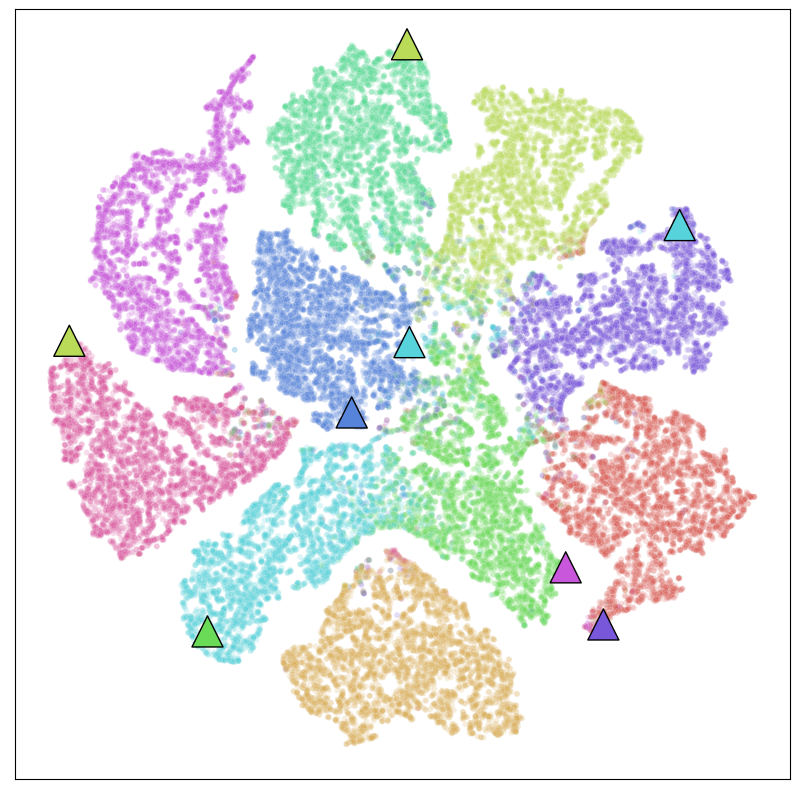}}
    \caption{t-SNE plots of CIFAR-10 datapoints in $\mathcal{D}_f$ (triangles) and $\mathcal{D}_r$ (dots) before and after unlearning. Colors indicate true labels for all plots. Regularization with adversarial examples and weight importance effectively preserves the decision boundary while migrating instances in $\mathcal{D}_f$ towards the class boundary to induce misclassification.}
    \label{fig:decision_boundary}
    \vspace{-2.0mm}
\end{figure}

\noindent\textbf{A1. Our method shows no pattern in misclassification.}   \ \
We first check whether the unlearned model classifies all instances in $\mathcal{D}_f$ to a particular set of labels. 
The model exhibiting no correlation between true labels and misclassified labels is crucial with respect to data privacy, as it indicates that the unlearning process avoids the so-called \textit{Streisand effect} where data instances being forgotten unintentionally becomes more noticeable~\cite{eternal}.
Figure~\ref{fig:confusion_matrix} shows the confusion matrices of (pre-unlearning label, post-unlearning label) pairs from $\mathcal{D}_f$ for $k=512$. We find no distinguishable pattern when unlearning with our methods as well as \textsc{NegGrad}, which shows that no specific label is used as a retainer, which adds another layer of security against adversaries in search of unlearned data points.

\noindent\textbf{A2. Our method effectively preserves the decision boundary.} \ \
We check whether the adversarial examples generated from $\mathcal{D}_f$ help in preserving the decision boundary in the feature space. Figure~\ref{fig:decision_boundary} shows t-SNE~\cite{(tsne)van2008visualizing} visualizations of final-layer activations from examples in $\mathcal{D}_r$ and $\mathcal{D}_f$ before and after unlearning. We find that unlearning through only negative gradient significantly distorts the previous decision boundary, leading to poor predictive performance across $\mathcal{D}_r$. However, when we incorporate adversarial samples from instances in $\mathcal{D}_f$, the decision boundary is well-preserved with unlearned examples being inferred as boundary cases in-between multiple classes. Even for examples that lie far from the decision boundary before unlearning, our method successfully relocates the corresponding representations towards the decision boundary, while keeping each class cluster intact.

\begin{figure}[!t]
    \centering
    \subfigure
    {\includegraphics[width=0.32\linewidth]{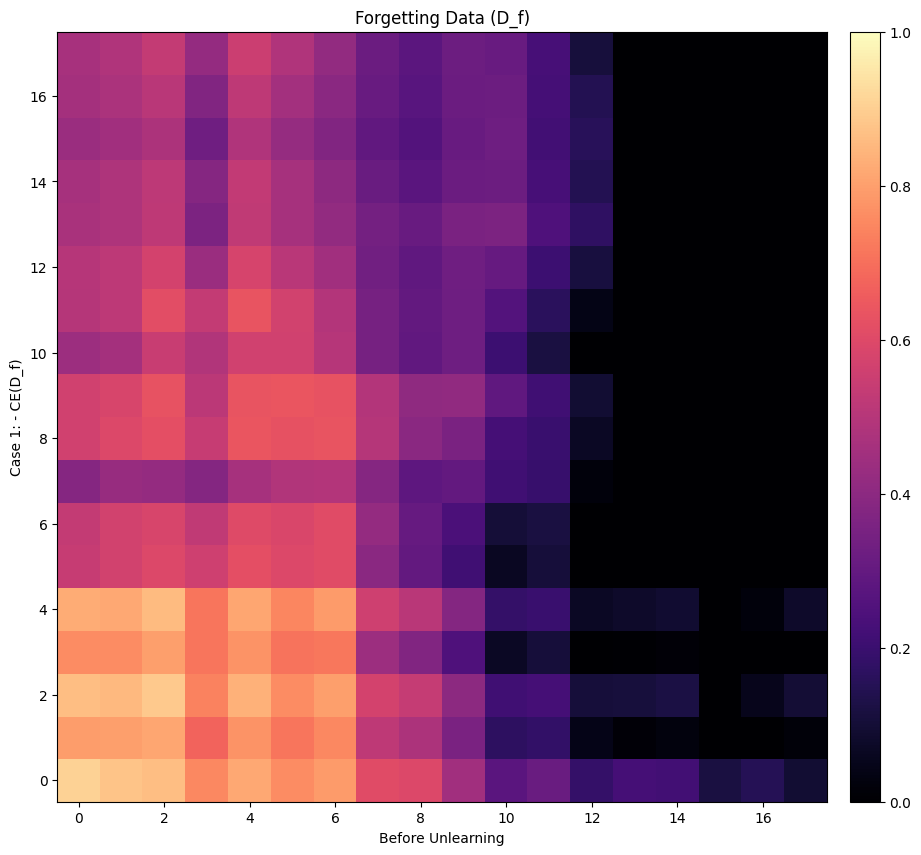}}
    \subfigure
    {\includegraphics[width=0.32\linewidth]{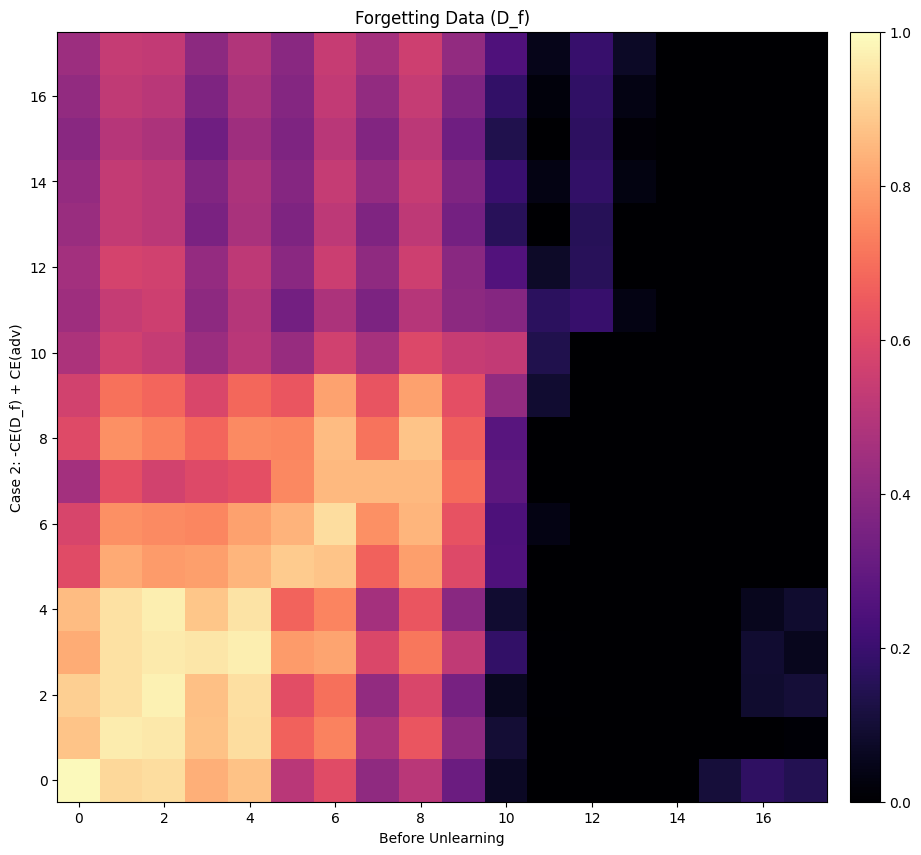}}
    \subfigure
    {\includegraphics[width=0.32\linewidth]{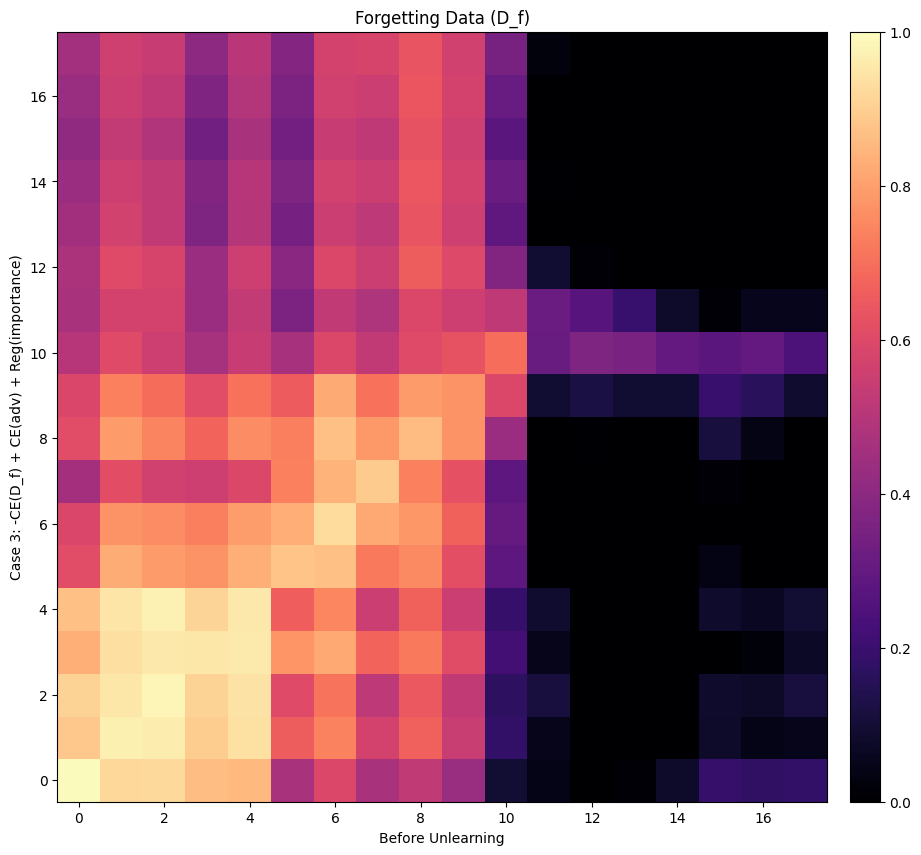}}
    \addtocounter{subfigure}{-3} 
    \subfigure[\textsc{NegGrad}]
    {\includegraphics[width=0.32\linewidth]{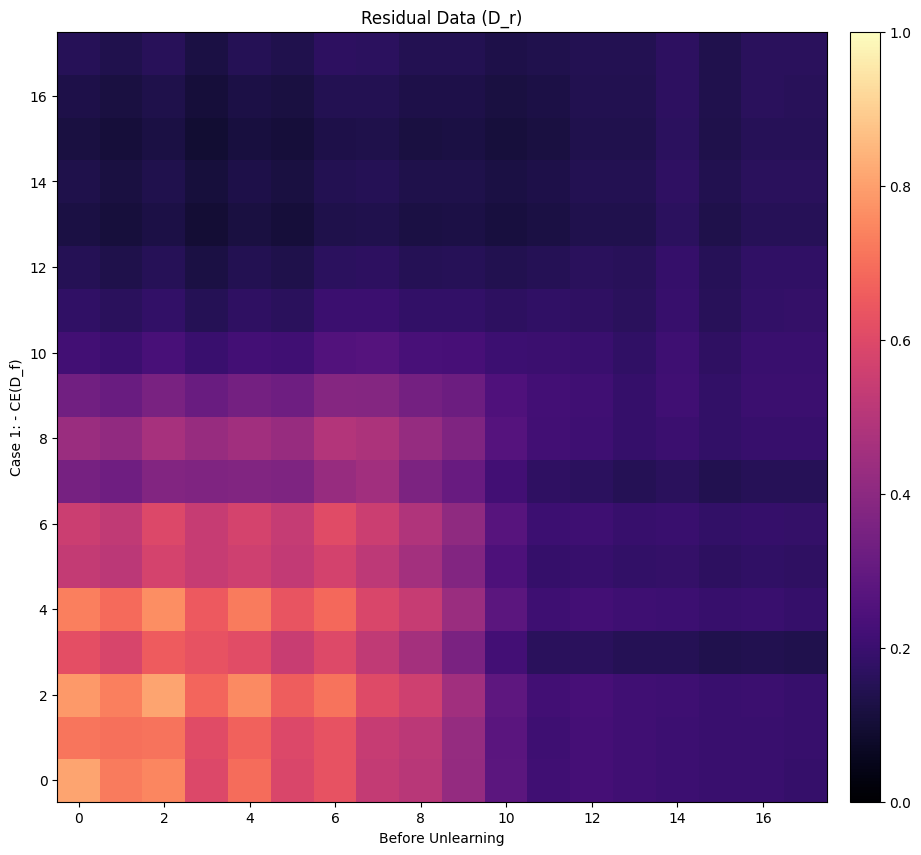}}
    \subfigure[\textsc{Adv}]
    {\includegraphics[width=0.32\linewidth]{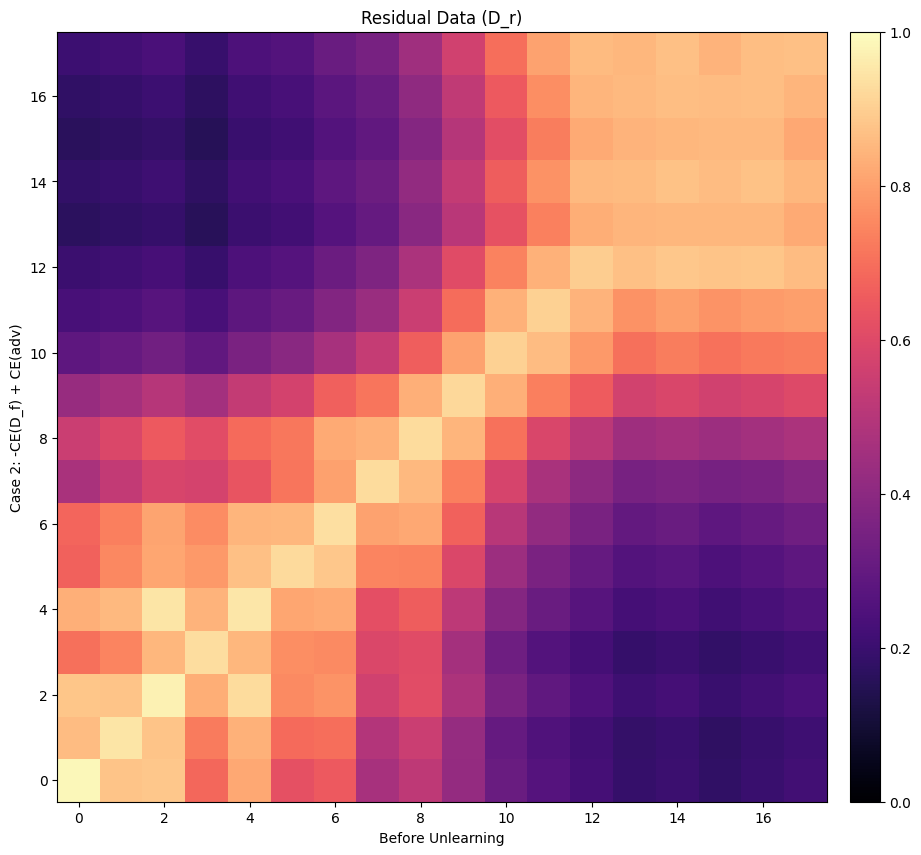}}
    \subfigure[\textsc{Adv+Imp}]
    {\includegraphics[width=0.32\linewidth]{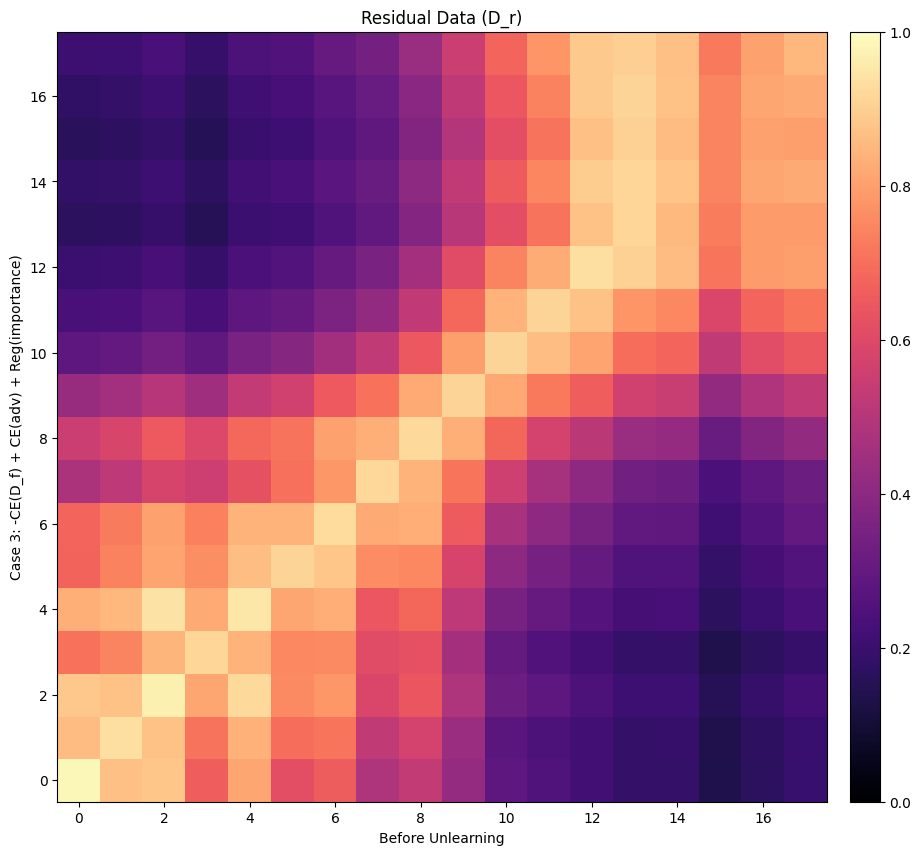}}
    \caption{Layer-wise CKA correlations on $\mathcal{D}_f$ (top row) and $\mathcal{D}_r$ (bottom row) between representations before (X-axis) and after (Y-axis) unlearning. Brighter color indicates higher CKA correlation. \textsc{NegGrad} results in large forgetting of high-level features in both $\mathcal{D}_f$ and $\mathcal{D}_r$. Our approaches selectively forget high-level features only in $\mathcal{D}_f$.}\label{fig:cka}
\end{figure}

\noindent\textbf{A3. Our method unlearns data by forgetting high-level features.} \ \  
Lastly, we compare the representations at each layer of the model before and after unlearning to identify where the intended forgetting occurs. For this analysis, we leverage CKA~\cite{(cka)kornblith2019similarity} that measures correlations between representations given two distinct models. Figure~\ref{fig:cka} shows the CKA correlation heatmaps between the original pre-trained model and the model after unlearning. Results show that for examples in $\mathcal{D}_f$, representations are no longer aligned starting from the 10-th layer, while representations before that layer still resemble those from the original model. This indicates that the model forgets examples by forgetting high-level features, while similarly recognizing low-level features in images as the original model. This insight is consistent with previous observations in the continual learning literature that more forgettable examples exhibit peculiarities in high-level features~\cite{toneva2018empirical}.

\section{Concluding Remarks}

We propose an instance-wise unlearning framework that deletes information from a pre-trained model given a set of data instances with mixed labels. Rather than undoing the influence of given instances during the pre-training, we aim for a stronger goal of unlearning via misclassifying or relabeling. We develop two regularization techniques that reduce forgetting on remaining data, one utilizing adversarial examples of deleting instances and another leveraging weight importance to focus updates on parameters responsible for propagating information we wish to forget. Both approaches are agnostic to the choice of architecture and require access only to the pre-trained model and instances requested for deletion. Experiments on various image classification datasets showed that our methods effectively mitigate forgetting on remaining data, while completely misclassifying the deleting data. 

\section{Acknowledgments}

This work was done while Sungmin Cha did a research internship at Advanced Machine Learning Lab, LG AI Research.

\appendix

\section{Pseudocode of our unlearning pipelines}

In this section, we provide pseudocode of our instance-wise unlearning framework. Algorithms~\ref{alg:pipeline_mis} and \ref{alg:pipeline_cor} show the details of two frameworks, one for unlearning via misclassification and another for unlearning via relabeling. The two pipelines are identical, except for the condition at which unlearning ends: when all examples in $\mathcal{D}_f$ are misclassified vs. when all examples in $\mathcal{D}_f$ are classified correctly to their new labels.

    \begin{algorithm}[H]
        \caption{Generate adversarial examples}
        \label{alg:generate_adversary}
        \begin{algorithmic}[1]
            \REQUIRE Forgetting data $\mathcal{D}_f$, Model $g_{\bm\theta}$\\
            \ENSURE Adversarial examples $\bar{\mathcal{D}}_f$
            \STATE $\bar{\mathcal{D}}_f \gets \emptyset$ 
            \FOR{$i$ \textbf{in range} $N_f$}
                \STATE $(\bm x^{(i)}, \bm y^{(i)}) \sim \mathcal{D}_f$
                \STATE Randomly sample $\bar{\bm y}^{(i)} \neq {\bm y}^{(i)}$
                \FOR{$j$ \textbf{in range} $N_{adv}$}
                \STATE $\bm {x}'^{(j)}_f \gets$ $L_2$-PGD$(\bm x^{(i)}, \bar{\bm y}^{(i)})$ (Eq.~\ref{eq:adversarial_attack})
                \STATE $\bar{\mathcal{D}}_f \gets \bar{\mathcal{D}}_f \cup \{(\bm {x}'^{(j)}_f, \bar{\bm y}^{(j)})\}$
                \ENDFOR
            \ENDFOR
            \STATE \textbf{return} $\bar{\mathcal{D}}_f$
        \end{algorithmic}
    \end{algorithm}

    \begin{algorithm}[H]
        \caption{Measure weight importance}
        \label{alg:weight_importance}
        \begin{algorithmic}[1]
            \REQUIRE Forgetting data $\mathcal{D}_f$, Model $g_{\bm \theta}$
            \ENSURE Weight importance $\bar{\Omega}$
            \vspace{1.5mm}
            \STATE $\bar{\Omega}\gets$ \{0\}
            \vspace{0.8mm}
            \STATE ${\Omega}\gets$ weight importances$(\mathcal{D}_f, g_{\bm \theta})$ (Eq.~\ref{eq:mas})
            \vspace{0.8mm}
            \FOR{$l$ \textbf{in range} $L$}
                \vspace{1.0mm}
                \STATE Get importance of $l$-th layer ${\Omega}^l \gets \Omega$
                \vspace{0.9mm}
                \STATE Normalize ${\Omega}^l \gets \dfrac{{\Omega}^l - Min({\Omega}^l)}{Max({\Omega}^l) - Min({\Omega}^l)}$
                \vspace{1.0mm}
                \STATE Update $\bar{\Omega}^{l} \gets \{1 - {\Omega}^l\}$
                \vspace{0.8mm}
            \ENDFOR
            \vspace{0.2mm}
            \STATE \textbf{return} $\bar{\Omega}$
        \end{algorithmic}
    \end{algorithm}

\begin{algorithm}
    \caption{Unlearning via misclassification (i.e., $\mathcal{L}_{\textrm{UL}}^{\textrm{MS}}$).}
    \label{alg:pipeline_mis}
    \begin{algorithmic}[1]
        \STATE UNLEARNACC = 100
        \STATE MAXEP = 100
        \STATE EP = 0
        \STATE $\bar{\mathcal{D}}_r \gets $ Generate adversarial examples with Algorithm 1
        \STATE $\bar{\Omega}\gets $ Measure weight importance with Algorithm 2
        \STATE $\Tilde{\bm \theta} \gets \bm \theta$ \ \ // \ \ Set the pretrained model for unlearning
        \WHILE{UNLEARNACC $\neq 0$}
        \STATE Minimize Eqn (6) or (7)
        \STATE UNLEARNACC = GetAccuracy($\mathcal{D}_f$, $g_{\Tilde{\bm \theta}}$)
        \IF{EP $>$ MAXEP}
            \STATE break
        \STATE EP += 1
        \ENDIF
        \ENDWHILE
        \STATE \textbf{return} $\tilde{\bm \theta}$
    \end{algorithmic}
\end{algorithm}

\begin{algorithm}
    \caption{Unlearning via relabeling (i.e., $\mathcal{L}_{\textrm{UL}}^{\textrm{Cor}}$).}
    \label{alg:pipeline_cor}
    \begin{algorithmic}[1]
        \STATE UNLEARNACC = 0
        \STATE MAXEP = 100
        \STATE EP = 0
        \STATE $\bar{\mathcal{D}}_r \gets $ Generate adversarial examples with Algorithm 1
        \STATE $\bar{\Omega}\gets $ Measure weight importance with Algorithm 2
        \STATE $\Tilde{\bm \theta} \gets \bm \theta$ \ \ // \ \ Set the pretrained model for unlearning
        \WHILE{UNLEARNACC $\neq 100$}
        \STATE Minimize Eqn (6) or (7)
        \STATE UNLEARNACC = GetAccuracy($\mathcal{D}_f$, $g_{\Tilde{\bm \theta}}$)
        \IF{EP $>$ MAXEP}
            \STATE break
        \STATE EP += 1
        \ENDIF
        \ENDWHILE
        \STATE \textbf{return} $\tilde{\bm \theta}$
    \end{algorithmic}
\end{algorithm}

\begin{algorithm}
    \caption{Unlearning via misclassification with RAWP.}
    \label{alg:pipeline_rawp}
    \begin{algorithmic}[1]
        \STATE UNLEARNACC = 100
        \STATE MAXEP = 100
        \STATE EP = 0
        \STATE $\Tilde{\bm \theta} \gets \bm \theta$ \ \ // \ \ Set the pretrained model for unlearning
        \STATE $\hat{\bm \theta} \gets \bm \theta$ \ \ // \ \ Set the pretrained model for AWP
        \WHILE{UNLEARNACC $\neq 0$}
        \STATE $\hat{\bm \theta}_{p} \gets \bm \hat{\bm \theta}$ \ \ // Set a proxy model
        \STATE Minimize $\mathcal{L}_{\textrm{UL}}^{\textrm{MS}}$ for $\hat{\bm \theta}_{p}$
        \STATE Get gradients and update $\hat{\bm \theta}_{p}$
        \STATE Get difference $\hat{\bm \theta}_{d}$ of each layer $i$ following \\ : $\hat{\bm \theta}_{d,i} = ||\Tilde{\bm \theta}_{i}||_{2} / (||\Tilde{\bm \theta}_{i} - \hat{\bm \theta}_{p,i}||_{2} + \epsilon) * (\Tilde{\bm \theta}_{i} - \hat{\bm \theta}_{p,i})$
        \STATE Perturb $\Tilde{\bm \theta}$ of each layer $i$ following \\ : $\Tilde{\bm \theta}_i$ = $\Tilde{\bm \theta}_i + \gamma \hat{\bm \theta}_{d,i}$
        \STATE UNLEARNACC = GetAccuracy($\mathcal{D}_f$, $g_{\Tilde{\bm \theta}}$)
        \IF{EP $>$ MAXEP}
            \STATE break
        \STATE EP += 1
        \ENDIF
        \ENDWHILE
        \STATE \textbf{return} $\hat{\bm \theta}$
    \end{algorithmic}
\end{algorithm}

\begin{table*}[!t]
\footnotesize
\caption{Results analogous to Table 1 of the manuscript, but with unlearning via relabeling each image in $\mathcal{D}_f$ to an arbitrarily chosen class. We see a similar trend where \textsc{Correct} loses significant information on $\mathcal{D}_r$, while our proposed methods retain predictive performance on $\mathcal{D}_r$ as well as $\mathcal{D}_{test}$.}
\vspace{-.1in}  
\centering
\smallskip\noindent
\resizebox{\linewidth}{!}{
\begin{tabular}{ll|cccc|cccc|cccc}
\toprule
&  & \multicolumn{4}{c|}{CIFAR-10} & \multicolumn{4}{c|}{CIFAR-100} & \multicolumn{4}{c}{ImageNet-1K}\\ 
&  & {$k=16$} & {$k=64$} & $k=128$ & $k=256$  & {$k=16$} & {$k=64$} & $k=128$ & $k=256$    & {$k=16$} & {$k=64$} & $k=128$ & $k=256$   \\ \midrule
\multirow{6}{*}{$D_f$ ($\uparrow$)} 
& \textsc{Before} & {0.0} & {0.0} & {0.0} & 0.0 & {0.0} & {0.0} & {0.0} & 0.0 & 0.0 & 0.0 & 0.0 & 0.0  \\
& \textsc{Oracle}  & 100.0 & 100.0 & 100.0 & 77.34 & 100.0 & 100.0 & 100.0 & 79.92 & 100.0  & 100.0  & 99.48 & 95.83  \\ 
& \textsc{Correct}  & 100.0   & 100.0   & 100.0   & 100.0     & 100.0   & 100.0   & 99.84 & 99.84  & 100.0 & 100.0 & 100.0 & 100.0  \\
& \textsc{RAWP}  & 100.0  & 100.0 & 100.0 & 96.41 & 100.0 & 99.38  & 99.06 & 93.75 & 100.0 & 98.96 & 56.25 & 39.06  \\ 
& \textsc{Adv}  & 100.0   & 99.38   & 98.28   & 96.17   & 100.0   & 100.0   & 98.28   & 92.42  & 100.0 & 89.06 & 71.88 & 58.59\\
& \textsc{Adv+Imp}   & 100.0   & 53.75   & 50.16   & 59.06  & 86.25   & 20.63   & 15.16   & 9.69  & 100.0 & 9.38 & 3.91 & 2.86\\
\midrule
\multirow{6}{*}{$D_r$ ($\uparrow$)}   
& \textsc{Before} & {99.60} & {99.60} & 99.60 & 99.60 & {99.98} & {99.98} & 99.98 & 99.98 & 87.48 & 87.42 & 87.42 & 87.42  \\
& \textsc{Oracle}  &  99.73 & 99.94 & 99.90 & 95.81 & 99.90 & 99.97 & 99.79 & 99.50 & 83.93  & 83.18 & 84.97 & 83.61  \\ 
& \textsc{Correct}  & {11.75} & {12.33}  & 9.71  & 8.97 & {74.84} & {31.79} & 18.64 & 6.93 & 82.47 & 77.21 & 68.19 & 44.47\\
& \textsc{RAWP}  & 79.94  & 63.95 & 52.12 & 21.14 & 59.36 & 40.93  & 9.20 & 6.15 & 48.66 & 31.55 & 0.46 & 0.24  \\ 
& \textsc{Adv}  & {85.53} & {83.36} & 81.06 & 72.41 & \textbf{92.94} & {94.64} & 96.32 & \textbf{94.28} & 83.39 & \textbf{84.79} &\textbf{84.52} & \textbf{84.11}\\
& \textsc{Adv+Imp}   & {\textbf{91.15}} & {\textbf{94.76}} & \textbf{90.57} & \textbf{81.82} & 90.73 & {\textbf{96.68}} & \textbf{96.44} & 92.40 & \textbf{83.58} & {83.31}& 80.43 &  79.29\\
\midrule
\multirow{6}{*}{$D_{test}$ ($\uparrow$)} 
& \textsc{Before}  & {92.59} & {92.59} & 92.59 & 92.59 & {77.10} & {77.10} & 77.10 & 77.10 & 76.01 & 76.01 & 76.01 & 76.01\\
& \textsc{Oracle}  & 91.65 & 91.99 & 91.57 & 87.51 & 74.05 &  74.93 & 74.15 & 73.81 & 72.76 & 72.04 & 73.83 & 72.79  \\ 
& \textsc{Correct} & {11.79} & {12.16}  & 9.80  & 8.87 & {53.11} & {24.37} & 14.64 & 5.95  & 71.57 & 66.61 & 58.37 & 38.29\\
& \textsc{RAWP}  & 73.58 & 59.48 & 48.02 & 20.25 & 41.93 & 29.51  & 7.88 & 5.71 & 42.40 & 27.52 & 0.48 & 0.23  \\ 
& \textsc{Adv}   & {79.15} & {76.95} & 74.61 & 66.80 & \textbf{65.62} & {66.79} & 68.56 & \textbf{66.75} & 72.34 & \textbf{73.51} & \textbf{73.38} & \textbf{73.00}\\
& \textsc{Adv+Imp}  & {\textbf{84.24}} & {\textbf{86.92}} & \textbf{82.82} & \textbf{74.68} & 64.28 & {\textbf{69.15}} & \textbf{68.60} & 64.81 & \textbf{72.49} & {71.92}& {69.20} & 68.46\\
\bottomrule
\end{tabular}
}
\vspace{-4mm}
\label{table:main_result_dataset_target}
\end{table*}

\begin{table}[!h]
\caption{Experimental results for various $\gamma$ of RAWP.}
\label{table:gamma_rawp}
\centering
\smallskip\noindent
\resizebox{\linewidth}{!}{
\begin{tabular}{cc|cccc}
\toprule
 & & \multicolumn{4}{c}{CIFAR-10} \\
\multicolumn{2}{c|}{} & $k=16$ & $k=64$ & $k=128$ & $k=256$ \\ 
\midrule
\multicolumn{1}{c}{\begin{tabular}[c]{@{}c@{}}RAWP\\ ($\gamma = 0.001$)\end{tabular}} & \begin{tabular}[c]{@{}c@{}}$D_f$\\ $D_r$\\ $D_{test}$\end{tabular} & \begin{tabular}[c]{@{}c@{}}0.0\\ 88.31\\ 81.66\end{tabular} & \begin{tabular}[c]{@{}c@{}}0.0\\ 65.44\\ 60.90\end{tabular} & \begin{tabular}[c]{@{}c@{}}0.0\\ 42.10\\ 39.16\end{tabular}  & \begin{tabular}[c]{@{}c@{}}0.0\\ 14.81\\ 14.34\end{tabular}\\ 
\midrule
\multicolumn{1}{c}{\begin{tabular}[c]{@{}c@{}}RAWP\\ ($\gamma = 0.005$)\end{tabular}} & \begin{tabular}[c]{@{}c@{}}$D_f$\\ $D_r$\\ $D_{test}$\end{tabular} & \begin{tabular}[c]{@{}c@{}}0.0\\ 81.70\\ 75.80\end{tabular} & \begin{tabular}[c]{@{}c@{}}0.0\\ 55.09\\ 51.79\end{tabular} & \begin{tabular}[c]{@{}c@{}}0.0\\ 28.60\\ 27.21\end{tabular}  & \begin{tabular}[c]{@{}c@{}}2.89\\ 13.45\\ 13.11\end{tabular}\\ 
\midrule
\multicolumn{1}{c}{\begin{tabular}[c]{@{}c@{}}RAWP\\ ($\gamma = 0.01$)\end{tabular}}  & \begin{tabular}[c]{@{}c@{}}$D_f$\\ $D_r$\\ $D_{test}$\end{tabular} & \begin{tabular}[c]{@{}c@{}}0.0\\ 67.15\\ 63.10\end{tabular} & \begin{tabular}[c]{@{}c@{}}0.0\\ 37.33\\ 35.47\end{tabular} & \begin{tabular}[c]{@{}c@{}}4.06\\ 13.28\\ 12.91\end{tabular} & \begin{tabular}[c]{@{}c@{}}6.17\\ 9.12\\ 9.12\end{tabular}\\ 
\midrule
\multicolumn{1}{c}{\begin{tabular}[c]{@{}c@{}}RAWP\\ ($\gamma = 0.1$)\end{tabular}} & \begin{tabular}[c]{@{}c@{}}$D_f$\\ $D_r$\\ $D_{test}$\end{tabular} & \begin{tabular}[c]{@{}c@{}}0.9\\ 10.97\\ 11.13\end{tabular} & \begin{tabular}[c]{@{}c@{}}6.25\\ 9.16\\ 9.09\end{tabular}  & \begin{tabular}[c]{@{}c@{}}8.43\\ 10.00\\ 10.00\end{tabular} & \begin{tabular}[c]{@{}c@{}}9.61\\ 10.00\\ 10.00\end{tabular} \\
\bottomrule
\end{tabular}
}
\end{table}

\section{Additional experimental results}

\paragraph{Results on relabeling.}

Table~\ref{table:main_result_dataset_target} shows results analogous to Table 1 of the manuscript, but with the goal of relabeling data points in $\mathcal{D}_f$ to arbitrarily chosen labels rather than misclassifying. 
We find that a similar trend, 
{first, \textsc{Oracle} achieves superior accuracy in most cases but we observe that it suffers from the trade-off between $D_f$ and the remaining datasets when the size of $k$ become large (\textit{e.g.}, $k=256$).
Second, \textsc{RAWP} shows much better performance than \textsc{Correct} in CIFAR-10, however, not for other datasets.
Third, \textsc{Adv} attains significantly less forgetting in $\mathcal{D}_r$ and $\mathcal{D}_{test}$ compared to \textsc{Correct}, while succesfully relabeling all points in most cases.
On the other hand, while \textsc{Adv+Imp} show even less forgetting, it loses accuracy in relabeling $\mathcal{D}_f$, showing that regularization via weight importance focuses too much on retaining previous knowledge rather than adapting to corrections provided in $\mathcal{D}_f$.}
An intuitive explanation on why this occurs particularly in relabeling is that while misclassifying can be done easily by driving the input to its closest decision boundary, relabeling can be difficult if the new class is far from the original class in the representation space. 
The difficulty rises even more when the size of $\mathcal{D}_f$ is large, in which case more parameters in the network are discouraged from being updated during unlearning.

\begin{table}[!h]
    \caption{Comparison against \textsc{Amnesiac} unlearning~\citep{graves2021amnesiac}.}
    \label{table:amnesiac}
    \centering
    \smallskip\noindent
    \resizebox{\linewidth}{!}{
    \begin{tabular}{ll|cccc}
        \toprule
        &  & \multicolumn{4}{c}{CIFAR-100 }\\ 
        &  & {$k=16$} & {$k=64$}  & {$k=128$} & {$k=256$} \\ 
        \midrule
        \multirow{5}{*}{$D_f$ ($\downarrow$)}
        & \textsc{Before}   & 95.83 & 99.83 & 99.83 & 99.84\\ 
        & \textsc{NegGrad}  & 0.0 & 0.0 & 0.0 & 0.0 \\ 
        & \textsc{Amnesiac}  & \textit{64.58} & \textit{65.10} & \textit{48.44} & \textit{11.46}\\ 
        & \textsc{Adv}   & 0.0 & 0.0 & 0.0 & 0.0 \\ 
        & \textsc{Adv+Imp} &  0.0 & 0.0 & 6.25 & 13.02\\ 
        \midrule
        \multirow{5}{*}{$D_r$ ($\uparrow$)}  
        & \textsc{Before}  & 99.82	& 99.48	& 99.74	& 99.61\\ 
        & \textsc{NegGrad}  & 87.88	& 50.85	& 32.92 & 9.66\\ 
        & \textsc{Amnesiac}  & 99.07 & 74.72 & 	51.09 &	15.07\\ 
        & \textsc{Adv}  & 89.14 & 75.10 & 65.91 & 35.57\\ 
        & \textsc{Adv+Imp}   &  92.31 & 79.50 & 68.34 & 38.12\\ 
        \midrule
        \multirow{5}{*}{$D_{test}$ ($\uparrow$)} 
        & \textsc{Before}  & 83.88 & 83.88 & 83.88 & 83.88\\ 
        & \textsc{NegGrad}  & 64.74 & 39.97 & 26.78 & 9.47\\ 
        & \textsc{Amnesiac}  & 79.99 & 39.63 & 	42.75 & 39.05\\ 
        & \textsc{Adv}   & 65.95 & 55.78 & 46.74 & 25.15\\ 
        & \textsc{Adv+Imp}  & 70.52 & 60.88 & 50.34 & 29.90\\ 
        \bottomrule
    \end{tabular}
    }
\end{table}

\paragraph{Results with other architectures.} 
Figure~\ref{fig:various_model_cifar100} shows unlearning results on CIFAR-100, but with different model architectures. We find that our methods effectively preserve knowledge outside the forgetting data, resulting in up to 40\% boost in accuracy. NegGrad again outperforms our methods when $k=4$, but soon breaks down when unlearning more instances. Interestingly, SqueezeNet and MobileNetv2 suffer from larger forgetting in $\mathcal{D}_r$ and $\mathcal{D}_{test}$ than ResNet-50, possibly due to the width being narrower as previously investigated by~\cite{mirzadeh2022architecture}. ViT also suffers from large forgetting, an observation consistent with previous work which showed that ViT suffers more catastrophic forgetting compared to other CNN-based methods in continual learning due to Transformer architectures requiring large amounts of data.
We also evaluate the results of unlearning on ImageNet-1K with varying $k$ in Figure~\ref{fig:various_model_imagenet}.
Our proposed methods prevent forgetting knowledge about the rest data $\mathcal{D}_r$ better than NegGrad in all cases where k is greater than 8.
At the same time, the methods effectively delete information about $\mathcal{D}_f$.

\paragraph{Comparison against \textsc{Amnesiac} unlearning.} 
Table~\ref{table:amnesiac} shows the results before and after unlearning $k$ instances.
To compare \textsc{Amnesiac}~\cite{graves2021amnesiac} with our methods, we use ResNet-18 models pretrained on CIFAR-100.
Note that these models are sourced from the official code provided by \citep{graves2021amnesiac}.
\textsc{Amnesiac} is known that an effective method for unlearning a small number of $k$, without loss of performance through the removal of parameter updates for sensitive data.
However, as we already shown in Table 1 of the manuscript,  it's important to mention that \cite{graves2021amnesiac} doesn't employ both $\mathcal{D}_f$ and $\mathcal{D}_r$ for class-wise unlearning. 
Instead, they necessitate the preservation of model parameter updates from each batch containing sensitive data. However, this requirement might not be feasible for real-world application due to its impracticality.
Also, it has been proven that it is effective for \textit{class-wise unlearning}.

Aligned with the manuscript, we focus to demonstrate the efficiency of both our proposed method and \textsc{Amnesiac} in more difficult \textit{instance-wise unlearning}.
Overall, the table presents that \textsc{Amnesiac} has poor performance, making it difficult to achieve unlearning in the mixed labels setting, showing high accuracy of $\mathcal{D}_f$ after unlearning.
As $k$ increases, it suffers from a significant loss of accuracy on the remaining data $\mathcal{D}_r$. 
This result again demonstrates that, as mentioned in the paper~\cite{graves2021amnesiac}, \textsc{Amnesiac} needs fine-tuning with $\mathcal{D}_r$, as a post-processing, to regain performance after the amnesiac unlearning. 
On the contrary, our approach not only effectively enables the unlearning of $\mathcal{D}_f$, but also maintains the knowledge from $\mathcal{D}_r$ and $\mathcal{D}_f$ when compared to the \textsc{Amnesiac} unlearning approach. Note that this holds true even as the value of $k$ increases, and all of this is achieved without requiring extra information like model parameter updates.


\begin{table}[!t]
    \centering
    \noindent
    \resizebox{\linewidth}{!}{
    \begin{tabular}{ll|ccccc}
        \toprule
        &  & \multicolumn{5}{c}{UTKFace}\\ 
        &  & $k=1$ & $k=2$  & $k=4$ & $k=8$ & $k=16$ \\ 
        \midrule
        \multirow{4}{*}{$D_f$ ($\downarrow$)}
        & \textsc{Before} & 100.00 & 100.00 & 100.00 & 100.00 & 100.00\\
        & \textsc{NegGrad} & 0.00 & 0.00 & 0.00 & 0.00 & 0.00\\
        & \textsc{Adv} & 0.00 & 0.00 & 0.00 & 0.00 & 0.00\\
        & \textsc{Adv+Imp} & 0.00 & 0.00 & 0.00 & 0.00 & 0.00\\
        \midrule
        \multirow{4}{*}{$D_r$ ($\uparrow$)}  
        & \textsc{Before} & 99.85 & 99.85 & 99.84 & 99.84 & 99.83\\
        & \textsc{NegGrad} & 75.75 & 36.80 & 28.95 & 28.35 & 32.75\\
        & \textsc{Adv} & 79.20 & 62.69 & 80.63 & 72.94 & 56.17\\
        & \textsc{Adv+Imp} & \bf 87.07 & \bf 63.85 & \bf 82.30 & \bf 74.33 & \bf 57.21\\
        \midrule
        \multirow{4}{*}{$D_{test}$ ($\uparrow$)} 
        & \textsc{Before} & 87.14 & 87.14 & 87.14 & 87.14 & 87.14\\
        & \textsc{NegGrad} & 72.67 & 36.57 & 29.40 & 27.16 & 32.14\\
        & \textsc{Adv} & 76.30 & 58.46 & 78.28 & 65.50 & 53.65\\
        & \textsc{Adv+Imp} & \bf 82.24 & \bf 61.79 & \bf 79.54 & \bf 66.98 & \bf 54.37\\
        \bottomrule
    \end{tabular}
    }
    \caption{Instance-wise unlearning results from a ResNet-18 age classifier trained on UTKFace.}\label{table:age_prediction}
\end{table}
    
\paragraph{Results from unlearning age classifiers on UTKFace.} For this experiment, we mirror a real-world setting and empirically test unlearning on a classifier that was trained on facial data. Specifically, we first pretrain ResNet-18 on age-group classification on the UTKFace dataset~\cite{(utkface)zhifei2017cvpr}. UTKFace is a dataset consisted of 20k facial images, each labeled with the age of the subject ranging between 0 and 116. Following previous work~\citet{(mfd)jung2021fair}, we divide the age range into three (0 to 19, 20 to 40, and more than 40), effectively reducing the task into 3-way classification. After pretraining, we apply instance-wise unlearning to adjust model parameters towards misclassifying $k$ facial images, while aiming to retain its predictive performance on the remaining images. 

Table~\ref{table:age_prediction} shows the unlearning results. For \textsc{NegGrad}, we find that its performance degrades quickly, with its classification performance being lower than that of a random baseline (33\% since task is 3-way classification) when unlearning more than 4 instances. On the other hand, both \textsc{Adv} and \textsc{Adv+Imp} consistently outperform the na\"ive baseline across all $k$, showing that our regularization techniques are also effective in the facial image domain. The gap between \textsc{Adv} and \textsc{Adv+Imp} is much larger with small $k$, which demonstrates high potential of weight importance-based regularization for unlearning under a continuous stream of instance-wise data deletion requests.

\section{Details of RAWP}

Algorithm \ref{alg:pipeline_rawp} presents the pseudocode of RAWP. As an alternative to minimizing cross-entropy loss, we employ adversarial weight perturbation (AWP)~\cite{(awp)wu2020adversarial} for our unlearning framework. The main difference between AWP and cross-entropy loss is that AWP does not directly minimize loss through gradient descent, but instead introduces a worst-case weight perturbation with a relative perturbation size constrained by $\gamma$. In the manuscript, we reported experimental results using the default value of $\gamma=0.01$.

Table \ref{table:gamma_rawp} reports additional experimental results from applying RAWP for unlearning a ResNet-18 pretrained on CIFAR-10 using various $\gamma$. We confirm that under various hyperparameterization for $\gamma$, RAWP consistently shows performance degradation, especially for large $k$.

\bibliography{aaai24}

\end{document}